\documentclass[final]{cvpr}

\usepackage{times}
\usepackage{epsfig}
\usepackage{graphicx}
\usepackage{amsmath}
\usepackage{amssymb}

\usepackage{enumitem}
\usepackage{booktabs}
\usepackage{tabulary}
\usepackage{makecell}
\usepackage{mathtools}
\usepackage{acronym}
\usepackage{caption}
\usepackage{array}
\usepackage{subcaption}
\usepackage{booktabs}
\usepackage{multicol}
\usepackage{multirow}
\usepackage[percent]{overpic}
\usepackage{tabularx} 
\renewcommand{\arraystretch}{1.1}
\newcolumntype{Y}{>{\centering\arraybackslash}X}

\usepackage[pagebackref=true,breaklinks=true,colorlinks,bookmarks=false]{hyperref}

\begin{document}

%%%%%%%%% TITLE
\title{
CoCosNet v2: Full-Resolution Correspondence Learning for Image Translation
\vspace{-1em}
}

\author{
Xingran Zhou$^1$\thanks{Author did this work during his internship at Microsoft Research Asia.}
\quad
Bo Zhang$^2$\quad
Ting Zhang$^2$\quad 
Pan Zhang$^4$ \quad
Jianmin Bao$^2$\quad \\
Dong Chen$^2$\quad
Zhongfei Zhang$^3$\quad 
Fang Wen$^2$
\vspace{+0.3em}
\\
$^1$Zhejiang University\quad$^2$Microsoft Research Asia\quad
$^3$Binghamton University\quad$^4$USTC
}

\maketitle
\thispagestyle{empty}
\pagestyle{empty}

%%%%%%%%% ABSTRACT
\begin{abstract}
	 We present the full-resolution correspondence learning for cross-domain images, which aids image translation. We adopt a hierarchical strategy that uses the correspondence from coarse level to guide the fine levels. At each hierarchy, the correspondence can be efficiently computed via PatchMatch that iteratively leverages the matchings from the neighborhood. Within each PatchMatch iteration, the ConvGRU module is employed to refine the current correspondence considering not only the matchings of larger context but also the historic estimates. The proposed CoCosNet v2, a GRU-assisted PatchMatch approach, is fully differentiable and highly efficient. 
% The proposed GRU-assisted PatchMatch approach is fully differentiable and highly efficient. 
	When jointly trained with image translation, full-resolution semantic correspondence can be established in an unsupervised manner, which in turn facilitates the exemplar-based image translation. Experiments on diverse translation tasks show that CoCosNet v2 performs considerably better than state-of-the-art literature on producing high-resolution images.
\end{abstract}
\vspace{-1em}

%%%%%%%%% BODY TEXT
\section{Introduction}

Image-to-image translation learns the mapping between image domains and has shown success in a wide range of applications~\cite{liu2017unsupervised,choi2018stargan,murez2018image,wan2020bringing,zhou2019text}.
Particularly, exemplar based image translation allows more flexible user control by conditioning the translation on a specific exemplar with the desired style. However, simultaneously producing high quality while being faithful to the exemplar is non-trivial, whereas it becomes rather challenging for producing high-resolution images.

Early studies~\cite{chen2017photographic,isola2017image,zhang2017stackgan,zhang2019self,wang2018high,brock2018large} directly learn the mapping through generative adversarial networks~\cite{goodfellow2014generative,mirza2014conditional}, yet they fail to leverage the information in the exemplar. Later, a series of methods~\cite{dumoulin2016learned, huang2017arbitrary,park2019semantic} propose to refer to the exemplar image during the translation, by modulating the feature normalization according to the style of the exemplar image. However, as the modulation is applied uniformly, only the global style can be transferred whereas the detailed textures are washed out in the final output.

\begin{figure}[t]
\centering
\includegraphics[width=0.97\columnwidth]{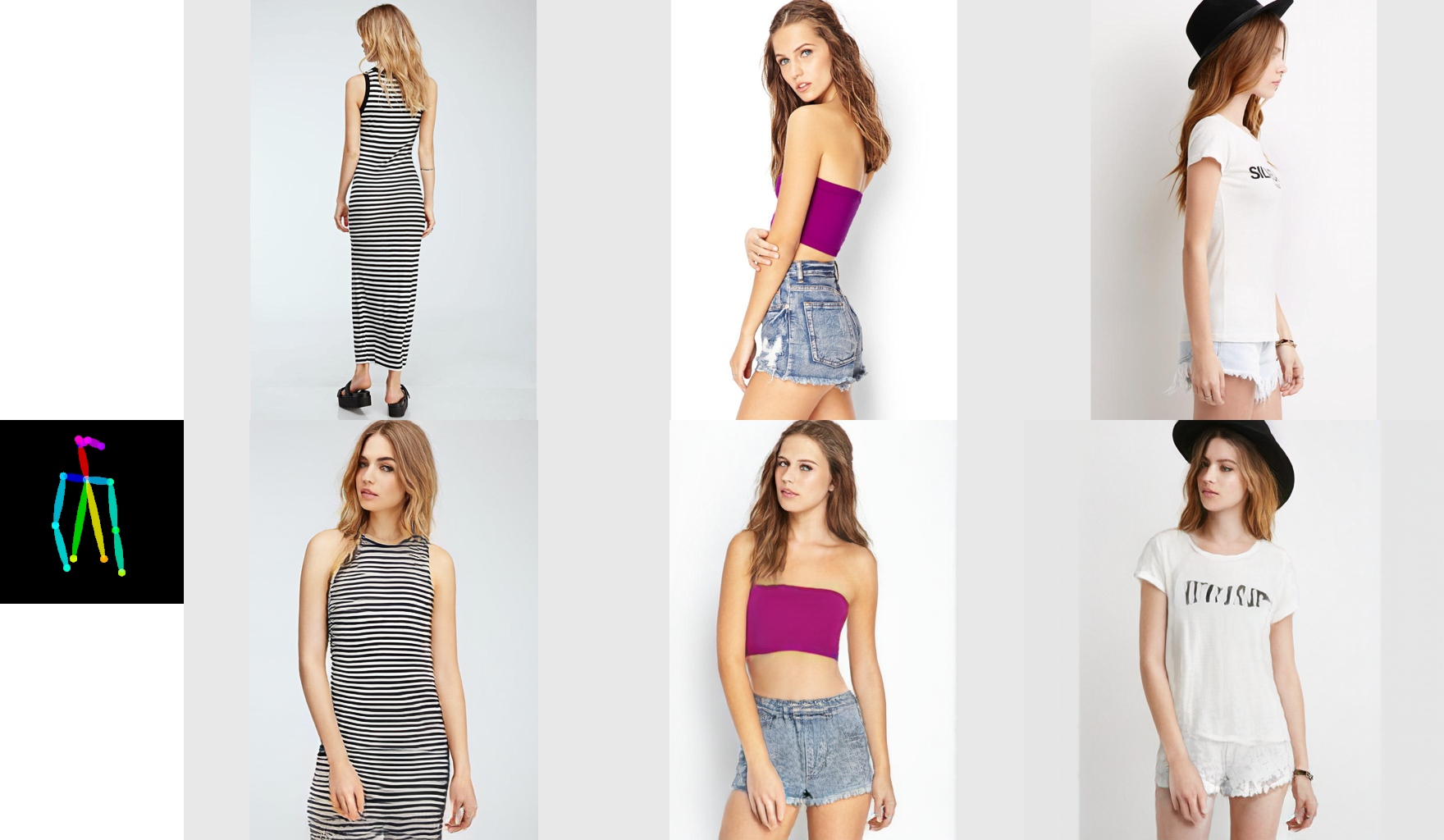}\\
\includegraphics[width=0.97\columnwidth]{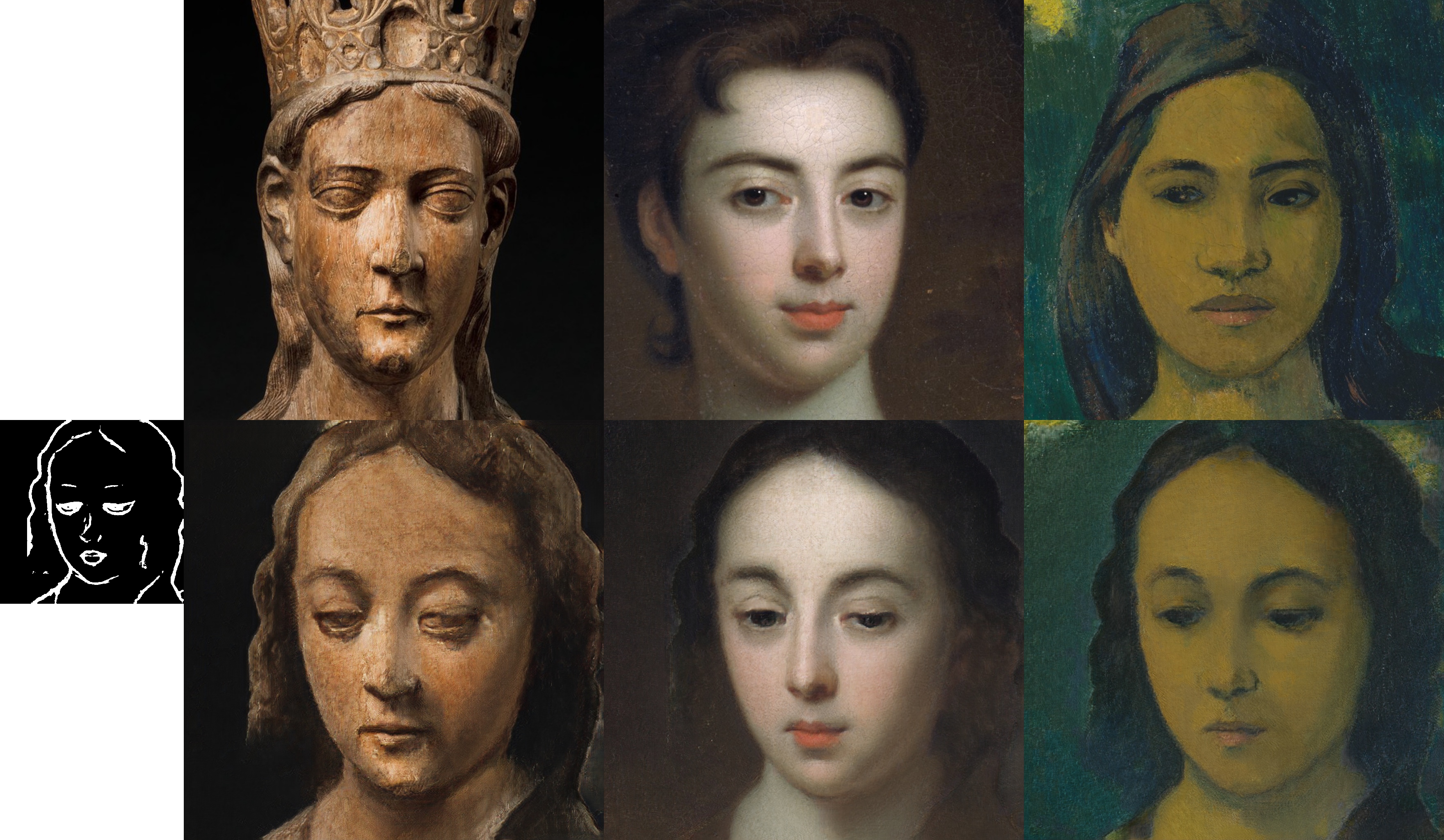}
\caption{Image translation at resolution 512$\times$512 for pose-to-body (DeepFashion) and at resolution 1024$\times$1024 for edge-to-face (MetFaces). For each task, the 1st row shows the exemplar images, and the 2nd row shows the translation outputs.}
\label{figure:teaser}
\vspace{-4mm}
\end{figure}

Very recently, CoCosNet~\cite{zhang2020cross} established the dense semantic correspondence between cross-domain images. In this way the network could make use of the fine textures from the exemplar, which eases the hallucination for the local textures. However, prohibitive memory footprint occurs when estimating high-resolution correspondence, as the matching requires to compute the pairwise similarities among all locations of the input feature maps, while low-resolution correspondences (\eg, 64$\times$64) cannot guide the network to leverage the fine structures from the exemplar. 

In this paper, we propose the cross-domain correspondence learning, \emph{in full-resolution} for the first time, which leads to high-resolution translated images in photo-realistic quality, as the network can leverage the meticulous details from the exemplar. To achieve that, we draw inspiration from PatchMatch~\cite{barnes2009patchmatch} which is advantageous in computational efficiency and texture coherency as it iteratively propagates the correspondence from the neighborhood rather than searching globally. Nonetheless, directly applying PatchMatch to high-resolution feature maps for training is infeasible and the reasons are threefold. First of all, this algorithm is not efficient enough for high-resolution images when the correspondence is initialized randomly. Second, at the early training phase, the correspondence is chaotic and the backward gradient flows to the incorrectly corresponded patches, making the feature learning difficult. Moreover, PatchMatch fails to consider a larger context when propagating the correspondence estimate and requires a large number of iterations to converge. 

To tackle these limitations, we propose the following techniques to learn the full-resolution correspondence. 1) We adopt a hierarchical strategy that makes use of the matchings from the coarse level to guide the subsequent, finer levels so that the searching at the fine levels may start with a good initialization. 2) Enlightened by the recent success of recurrent refinement~\cite{ren2019progressive,cai2019disentangled,teed2020raft}, we employ convolutional gated recurrent unit (ConvGRU) to refine the correspondence within each PatchMatch iteration. The GRU-assisted PatchMatch considers a larger context as well as the historic correspondence estimates, which considerably improves the correspondence quality. Besides, it greatly benefits the feature learning as the gradient can now flow to a larger context than just a few corresponded patches. 3) Last but not least, the proposed hierarchical GRU-assisted PatchMatch is fully differentiable, and learns the cross-domain correspondence in an unsupervised manner, which is very challenging especially in high-resolution.
%\ie, learning from the image warping rather than receiving direct supervision.

We show that our method, called CocosNet v2, achieves significantly higher quality images than the state-of-the-art litearture due to the full-resolution cross-domain correspondence. More importantly, our approach is able to generate visually appealing image translation results in high-resolution, \eg, images at 512$\times$512 and 1024$\times$1024 (Figure~\ref{figure:teaser}).
We summarize our major contributions as follows: 
\begin{itemize}[leftmargin=*,topsep=1pt]
    \itemsep-0.2em
	\item We propose to learn full-resolution correspondence from different domains in order to capture meticulously realistic details from an exemplar image for image translation.
	\item To achieve that, we propose CoCosNet v2, a hierarchical GRU-assisted PatchMatch method, for efficient correspondence computation, which is simultaneously learned with image translation.
	\item We show that the full-resolution correspondence leads to significantly finer textures in the translation output. The translated images demonstrate unprecedented quality at large resolutions.
\end{itemize}

%-------------------------------------------------------------------------
\section{Related works}

\noindent{\bf PatchMatch.}
Correspondence matching
is a fundamental problem in computer vision~\cite{brox2010large,liu2010sift,weinzaepfel2013deepflow,lucas1981iterative,duggal2019deeppruner,efros1999texture,wei2000fast}.
The prohibitively high computational challenge has been largely alleviated by the pioneering work, PatchMatch~\cite{barnes2009patchmatch}. The key insights stem from two principles:
1) good patch matches can be found via random sampling; 2) images are coherent such that matches can be propagated to nearby areas. Due to its efficiency, PatchMatch has been successfully applied to different tasks~\cite{li2015spm,bleyer2011patchmatch,bao2014fast,hu2016efficient,duggal2019deeppruner}. However, traditional PatchMatch can only find matches with image and is unsuitable to deep neural networks. Recently, ~\cite{duggal2019deeppruner} proposes to make the whole matching process differentiable and enables the feature learning and correspondence learning end-to-end. However, this method is still computational prohibitive to learn high-resolution correspondence during training. In contrast, we apply PatchMatch in hierarchy, and propose a novel GRU-assisted refinement module to consider a larger context, which enables a faster convergence and a more accurate correspondence. It is worth noting that~\cite{li2016combining,liao2017visual} use PatchMatch for style transfer, but they operate on the pretrained VGG features and require the input to be natural images, whereas we allow the feature learning for arbitrary domain inputs such as pose or edge. 

\begin{figure*}[t]
\centering
\includegraphics[width=1.98\columnwidth]{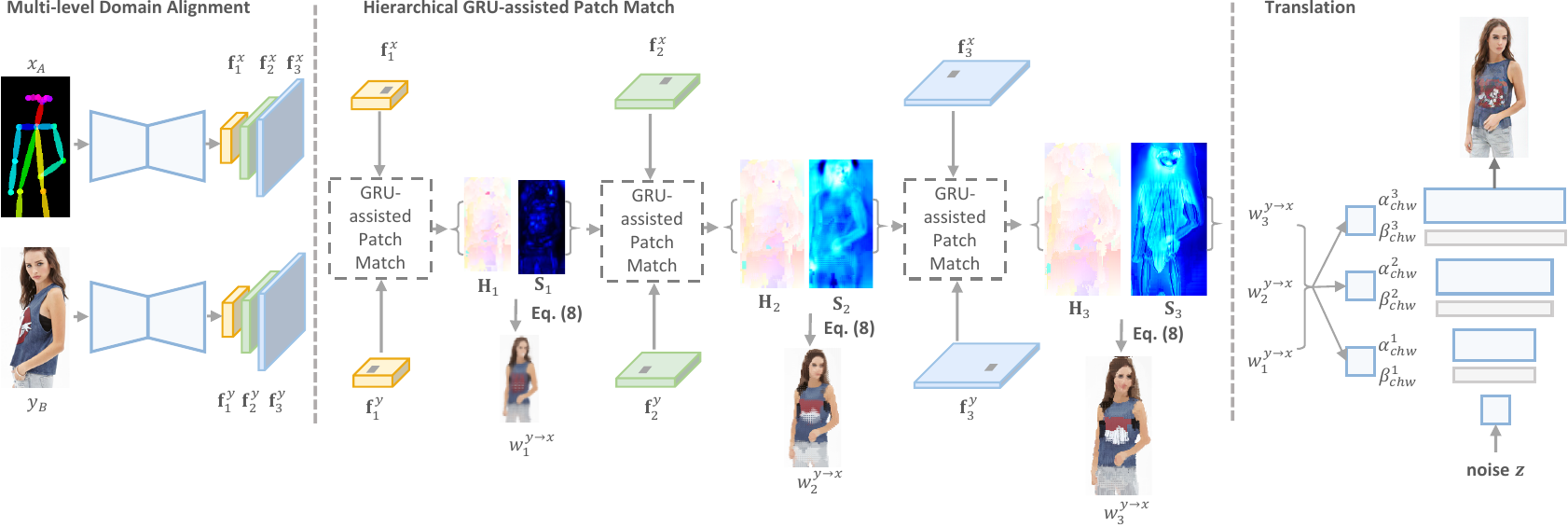}
\vspace{-0.5em}
\caption{The overall architecture of CoCosNet v2. We learn the \emph{cross-domain correspondence in full resolution}, by which we warp the exemplar images ($w_i^{y\to x}$) and feed them into the translation network for further rendering. The full-resolution correspondence is learned hierarchically, where the low-resolution result serves as the initialization for the next level. In each level, the correspondence can be efficiently computed via differentiable PatchMatch, followed by ConvGRU for recurrent refinement.}
\label{figure:framework}
\vspace{-6mm}
\end{figure*}

%-------------------------------------------------------------------------
\noindent{\bf Image-to-image translation.}
Image translation methods~\cite{isola2017image,wang2018high,park2019semantic,zhu2017unpaired,yi2017dualgan,kim2017learning,liu2017unsupervised,royer2020xgan} typically resort to a conditional generative adversarial network and optimize the network through either paired data with explicit supervision or unpaired data by enforcing cycle consistency. Recently, exemplar-based image translation~\cite{huang2018multimodal,qi2018semi,wang2019example,ma2018exemplar,riviere2019inspirational,bansal2019shapes,zhang2019deep} have attracted a lot of interest due to its flexibility and improved generation quality. While most methods transfer the global style from the reference image, a recent work, CoCosNet~\cite{zhang2020cross} proposes establishing the dense semantic correspondence to the cross-domain inputs, and thus better preserves the fine structures from the exemplar. Our work is closely related to CoCosNet~\cite{zhang2020cross} but has a substantial improvement.
We aim to compute dense correspondence on full-resolution whereas \cite{zhang2020cross} can only find the correspondence on a small scale. Due to the full-resolution correspondence, our network can leverage finer structures from the exemplar, and thus achieves a superior quality on high-resolution outputs.

\section{CoCosNet v2}

Given an image $x_A$ in the source domain $\mathcal{A}$ and an image $y_B$ in the target domain $\mathcal{B}$, we propose to learn full-resolution cross-domain correspondences that aim to capture finer details and serve as a better guidance in exemplar-based image translation.
Specifically, $x_A$ and $y_B$ are first represented as multi-level features (Section~\ref{sec:domainalignment}).
Thereafter the correspondences are established starting from low-resolution to full-resolution, which are further used to warp the exemplar to align with $x_A$ (Section~\ref{sec:approach}).
Finally, the warped exemplars are passed through a translation network to generate the desired output image (Section~\ref{sec:translation}).
We illustrate the whole network architecture in Figure~\ref{figure:framework}.

\subsection{Multi-level domain alignment} \label{sec:domainalignment}
We first learn a common latent space $\mathcal{S}$ in which the representation contains the semantic contents for both domains and the features can be compared under some similarity metric. 
Similar to prior work~\cite{zhang2020cross}, we learn two mapping functions for both domains respectively. We build a pyramid of $L$ latent spaces ranging from low-resolution to high-resolution,
instead of creating merely one latent space.
For feature extraction,
we adopt a U-net architecture to enable rich contextual information being propagated to higher resolution features by means of skip connections.   

Formally, let $\mathcal{M}_A$ and $\mathcal{M}_B$ be the corresponding two mapping functions, we have the multi-level latent features,
\begin{align}
\mathbf{f}^x_{1}, \cdots, \mathbf{f}^x_{L} & = \mathcal{M}_A (x_A; \theta_{\mathcal{M}_A}),\\
\mathbf{f}^y_{1}, \cdots, \mathbf{f}^y_{L} & = \mathcal{M}_B (y_B; \theta_{\mathcal{M}_B}),
\end{align}
where $\mathbf{f}^x_{l} \in \mathbb{R}^{H_lW_l\times C_l}$
with the height $H_1 < \cdots < H_L$, width $ W_1 < \cdots < W_L$, and $C_l$ denotes channel number. 
Latent features $\{\mathbf{f}^x_{1}, \cdots, \mathbf{f}^x_{L}\}$ are enlarged from small resolution to the full resolution. $\{\mathbf{f}^y_{1}, \cdots, \mathbf{f}^y_{L}\}$ have similar meanings, whereas, $ \theta_{\mathcal{M}_A}$ and $\theta_{\mathcal{M}_B}$ denote the parameters.

\subsection{Hierarchical GRU-assisted PatchMatch} \label{sec:approach}
It is worth noting the previous works compute dense correspondence field at the low-resolution level because of memory constraints and speed limitations.
We propose to exploit the correspondences on the full-resolution feature level, 
\ie, $\mathbf{f}^x_{L}$ and $\mathbf{f}^y_{L}$, 
and present a novel effective approach that is much less demanding in memory and time.

\noindent {\bf Coarse-to-fine strategy.}
Directly establishing the correspondences on full-resolution features not only increases the 
computational complexity, but also magnifies the noise and ambiguities of small patches. 
To deal with that, we propose a coarse-to-fine strategy on the pyramid of latent representations.
In particular, we start with correspondence matching at the lowest resolution level,
and use the matching results 
as the initial guidance 
at the subsequent, higher-resolution level.
In this way, the correspondence fields of all the levels 
can be acquired.
Formally we have,
\begin{equation}
\mathbf{H}_l = \mathcal{N}_l(\mathbf{H}_{l-1}, \mathbf{f}_l^x, \mathbf{f}_l^y),
\end{equation}
where $\mathbf{H}_l \in \mathbb{R}^{H_l W_l \times 2K}$ is the nearest neighbor field for $\mathbf{f}_l^x$.
Specifically, for a feature point $\mathbf{f}_l^x(\mathbf{p})$, $\mathbf{H}_l(\mathbf{p})$ specifies the locations of its top $K$ nearest neighbors in $\mathbf{f}_l^y$. %using some distance measure. 
We have
\begin{align}
\mathbf{H}_{l}(\mathbf{p},1) = \mathop{\arg \min}\limits_{\mathbf{q}} d(\mathbf{f}_l^x(\mathbf{p}), \mathbf{f}_l^y(\mathbf{q})), \label{eqn:warp}
% \vspace{-8em}
\end{align}
as an example.
Yet it takes a lot of time to traverse $\mathbf{p}$ and $\mathbf{q}$ exhaustively, especially on the entire full-resolution feature map. Therefore, we propose the GRU-assisted PatchMatch, which attempts an iterative improvement.

\begin{figure}[t]
\centering
% \begin{overpic}[scale=0.54]{figs-camera-ready/F2-crop1.pdf}
% \put(22,-2){\footnotesize (a)}
% \put(65,-2){\footnotesize (b)}
% \put(2,20){\footnotesize $\mathbf{H}_{l,0}$}
% \put(40,24){\footnotesize $\mathbf{H}'$}
% \put(80.8,19.5){\footnotesize $\mathbf{\Delta H}$}
% \put(94,20){\footnotesize $\mathbf{H}_{l}$}
% \put(46,51){\footnotesize $\mathbf{H}$}
% \put(62,42){\footnotesize $h$}
% \end{overpic}
 \includegraphics[width=0.5\textwidth]{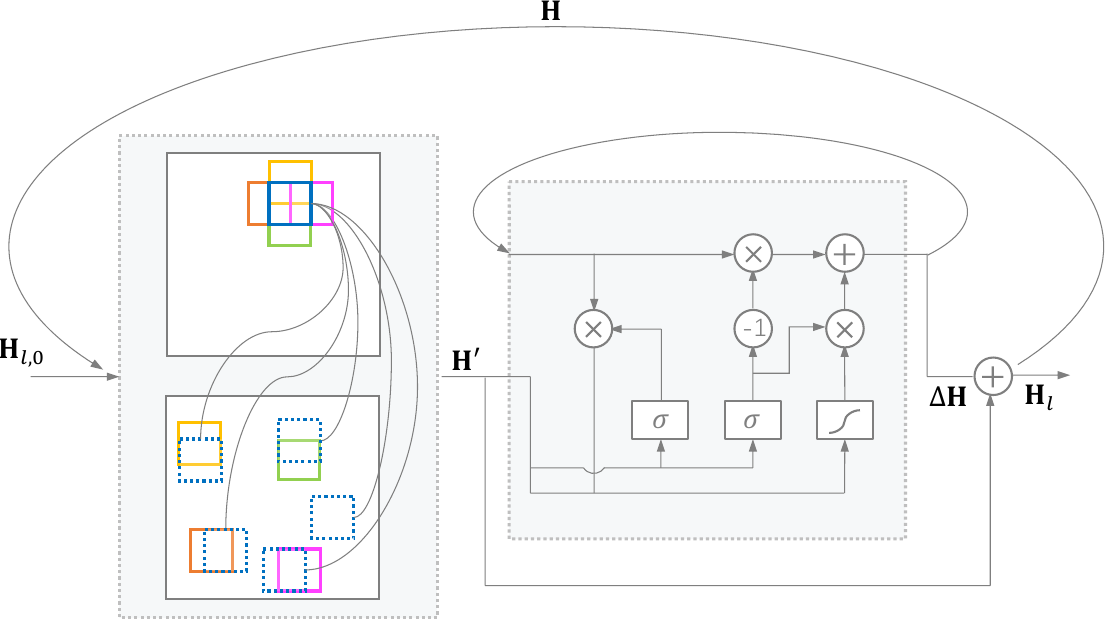}
\caption{GRU-assisted PatchMatch consisting of (a) propagation and (b) GRU-based refinement. Note that the propagation for all the locations are conducted in parallel.}
\label{figure:framework2}
\vspace{-6mm}
\end{figure}

\noindent {\bf GRU-assisted PatchMatch.}
Essentially, our algorithm can be briefly viewed as performing  \emph{propagation}
and \emph{GRU-based refinement} iteratively and recurrently
until convergence or a fixed number of iterations is reached. 
The previous level results $\mathbf{H}_{l-1}$ are utilized as the initialization,
and are improved gradually by alternating the two steps. 
We illustrate this matching process in Figure~\ref{figure:framework2}.

We denote the correspondence map in the $t$th step as $\mathbf{H}_{l,t}$, and the initialization correspondence field $\mathbf{H}_{l,0}$ is upsampled from $\mathbf{H}_{l-1}$.
The level annotation $l$ is omitted in this subsection without causing confusion.
The first step, propagation, stems from the seminal work PatchMatch~\cite{barnes2009patchmatch}.
It improves the matching of the current patch by 
examining the already known matching results of its neighborhoods, which we denote as, 

\begin{equation}
	\mathbf{H}'_t = propagation(\mathbf{H}_t, \mathbf{f}^x, \mathbf{f}^y).
\end{equation}
where $\mathbf{H}_{t}^{\prime}$ are the nearest neighbor field (NNF)  propagation results. 
% However, propagation only checks exactly one nearest neighbor, which makes it heavily rely on the spatial smoothness assumption and tend to be trapped in a local optimum.
However, propagation only checks spatially adjacent patches, which makes it heavily relying on the spatial smoothness assumption and tends to be trapped in a local optimum.
The random search step in PatchMatch does alleviate this issue to some degree, but it is not enough especially when searching in an extremely large candidate set.
Our solution is to look up distant candidates selectively rather than randomly searching, which is guided through a novelly designed refinement module. We expect that, given current offsets, the operator outputs a refinement field that serves as a correction to some incorrectly matched pairs.

Specifically in the second step, we adopt a convolutional gated recurrent unit (ConvGRU),
\begin{equation}
\vspace{-0.5em}
\begin{split}
	z_t &= \sigma(\text {Conv}([h_{t-1},x_t], \theta_z)) \\
	r_t &= \sigma(\text {Conv}([h_{t-1},x_t], \theta_r)) \\
	\hat{h}_t &= \text{tanh}(\text {Conv}([r_t \odot h_{t-1},x_t], \theta_h))\\
	h_t &= (1-z_t) \odot h_{t-1} + z_t \odot \hat{h}_t
\end{split}
\vspace{-0.2em}
\end{equation}
where $x_t$ is the input obtained by concatenating features extracted from four variables: $\mathbf{f}^x$, $\mathbf{f}^y$, $\mathbf{O}_t$, $\mathbf{S}_t$.
$\mathbf{O}_t$ and $\mathbf{S}_t$ are the current offset and the corresponding matching score,
\begin{equation}
\begin{split}
	\mathbf{O}_t(\mathbf{p},k) &= \mathbf{H}'_t(\mathbf{p},k) - \mathbf{p}, \\
	\mathbf{S}_t(\mathbf{p},k) &= \cos (\mathbf{f}^x(\mathbf{p}), \mathbf{f}^y(\mathbf{H}'_t(\mathbf{p},k)),
	\label{eqn:o_s}
\end{split}
\end{equation}
where $k=1,2,\cdots,K$ considering $K$ nearest neighbors. 
The initial hidden state is set as $\mathbf{0}$ and the offset update $\Delta \mathbf{H}_t$ is predicted by feeding the output hidden state $h_t$ to two convolutional layers.
At last, the offsets are updated by: $\mathbf{H}_{t+1} = \mathbf{H}'_t + \Delta \mathbf{H}_t$ and are passed to the next step.

\noindent\textbf{The benefits of ConvGRU.} First, it helps refine the current correspondence estimate making use of a larger context, rather than the local neighborhood. The correspondence can therefore become globally coherent with a faster convergence. Second, the GRU memorizes the history of correspondence estimate, and somehow forecasts the possible corresponding location in the next iteration. Third, the backward gradient can now flow to the pixels in a larger context, rather than at a specific location, which benefits the feature learning and in turn the correspondence. 

\noindent {\bf Differentiable warping function.}
Unlike conventional applications that directly push the learned correspondences towards ground truth, we do not have the offset ground truth in image-to-image translation.
Instead, we leverage the correspondence field in the subsequent translation network to generate high-quality outputs, which pushes the correspondence field to be accurate.
% which we assume in turn will push the correspondence field to be accurate.

We take the correspondence field to warp the exemplar image $y_B$ and use the warped image $w_l^{y\rightarrow x}$ to guide the translation network.
Usually,  $w_l^{y\rightarrow x}$ is obtained by using only the nearest match,
\ie, $w_l^{y\rightarrow x}(\mathbf{p}) = y_B (\mathbf{H}_l(\mathbf{p},1))$. However, the $\arg \min$ operation in Equation~\ref{eqn:warp} is not differentiable. 
Therefore, we propose to use the following soft warping which is the average of top $K$ possible warping:
\begin{align}
w_l^{y\rightarrow x}(\mathbf{p}) = \sum_{k=1}^{K} softmax(\mathbf{S}_l(\mathbf{p}, k)) y_B(\mathbf{H}_l(\mathbf{p},k)) ,
\end{align}
where $\mathbf{S}$ is the matching score defined in Equation~\ref{eqn:o_s}, 
indicating the semantic similarity.

\subsection{Translation network} \label{sec:translation}
The translation network $\mathcal{G}$ aims to synthesize an image $\hat{x}_B$ that is desired to respect the spatial semantic structure in $x_A$ while resembling the appearance of similar parts in $y_B$.
Similar to recent conditional generators~\cite{miyato2018cgans,zhang2019self,mescheder2018training}, we employ a simple and natural way that takes a constant code $z$ as input.
To preserve the semantic information of the warped exemplar images $w^{y\rightarrow x}_1, \cdots, w^{y\rightarrow x}_{L}$, we resort to spatially-adaptive denormalization
(SPADE)~\cite{park2019semantic} that learns the modulation parameters adaptively. 

Specifically, let the activation before the $i^{th}$ normalization layer be $T^i \in \mathbb{R}^{C_i \times H_i \times W_i}$.
we first concatenate the warped images in the channel dimension (upsampling is performed here when necessary).
The resulting concatenation is denoted as $\hat{w}^{y\rightarrow x} = [w^{y\rightarrow x}_{1}\uparrow, \cdots, w^{y\rightarrow x}_L] $ where $\uparrow$ indicates upsampling. Thereafter we project $\hat{w}^{y\rightarrow x}$ through two convolutional layers to produce the modulation parameters $\alpha_{h,w}^i$ and $\beta_{h,w}^i$ for style modulation,
\begin{align}
\alpha_{h,w}^i(\hat{w}^{y\rightarrow x}) \times \frac {T^i_{c,h,w} - \mu_{h,w}^i} {\sigma_{h,w}^i} + \beta_{h,w}^i(\hat{w}^{y\rightarrow x}),
\end{align}
where $\mu_{h,w}^i$ and $\sigma_{h,w}^i$ are calculated mean and standard deviation.
Finally, the translation result can be obtained by,
\begin{align}
	\hat{x}_B = \mathcal{G}(z, \hat{w}^{y\rightarrow x} ;\theta_{\mathcal{G}}),
\end{align}
where $\theta_{\mathcal{G}}$ denotes the network parameters.

\subsection{Loss functions}
Our approach is end-to-end differentiable and can be optimized through backpropagation to simultaneously learn the cross-domain correspondence and the desired output. Generally, it is easy to access the semantically aligned data pair $\{x_A, x_B\}$ in different domains, but does not necessarily have the access to the training triplets $\{x_A,y_B,x_B\}$ where $x_B$ shares a similar appearance with $y_B$ while resembling the semantics of $x_A$. Hence we construct the \emph{pseudo exemplar} $\tilde{y}_B=\mathcal{T}(x_B)$ from $x_B$ by applying geometric distortion, where $\mathcal{T}$ denotes the geometric augmentation. 

\noindent {\bf Domain alignment loss.} 
%For successful correspondence, the multi-level representation for $x_A$ and its pseudo exemplar $\tilde{y}_B$ must lie in the same space, therefore we enforce,
%\begin{equation} 
%	\mathcal{L}_{align} = \|\mathcal{M}_A (x_A; \theta_{\mathcal{M}_A}) - \mathcal{M}_B (\tilde{y}_B; \theta_{\mathcal{M}_B})\|_1.
%\end{equation} 
For successful correspondence, the multi-level representation for $x_A$ and its corresponding counterpart $x_B$ must lie in the same space, therefore we enforce,
\begin{equation} 
	\mathcal{L}_{align} = \|\mathcal{M}_A (x_A; \theta_{\mathcal{M}_A}) - \mathcal{M}_B (x_B; \theta_{\mathcal{M}_B})\|_1.
\end{equation} 

\noindent {\bf Correspondence loss.}
Still, with the pseudo pairs, the warping $w^{\tilde{y}\rightarrow x}$ should exactly be $x_B$. Thus we enforce the correspondence with,
\begin{align}
	\mathcal{L}_{corr} = \sum_{l} \|w^{\tilde{y}_B\rightarrow x_A}_l - x_B {\downarrow}\|_1,
\end{align}
where $\downarrow$ indicates down-sampling to match the size of $x_B$ to the warped image.

\noindent {\bf Mapping loss.}
We expect that the cross-domain inputs can be mapped from the latent representation to their corresponding counterparts in the target domain, which helps the semantics-preserving in the latent space,
\begin{align}
	\mathcal{L}_{map} &= \|\mathcal{R}(\mathcal{M}_A (x_A;\theta_{\mathcal{M}_A}))-x_B\|_1 \\
	& +\| \mathcal{R}(\mathcal{M}_B (y_B; \theta_{\mathcal{M}_B})) - y_B\|_1,
\end{align}
where $\mathcal{R}$ maps the features to images in the target domain. 

\noindent {\bf Translation loss.}
The translated output is desired to be semantically similar to the input with the appearance close to that of the exemplar. We propose two losses focusing on the two objectives respectively. One is the perceptual loss to minimize the semantic discrepancy against $x_B$:
\begin{align}
	\mathcal{L}_{sem} = \|\phi_m(\hat{x}_B) - \phi_m(x_B) \|_1,
\end{align}
where we adopt features $\phi_m$ from high-level 
layers of pretrained VGG network.
The other one is the appearance loss that comprises of a contextual loss (CX)~\cite{mechrez2018contextual} when applying an arbitrary exemplar $y_B$ and a feature matching loss when using a pseudo exemplar $\tilde{y}_B$. The appearance loss encourages the appearance resemblance by leveraging low-level features $\phi_m$ of VGG. Concretely, the appearance loss is,  
\begin{align}
	\mathcal{L}_{app} &= \sum\limits_{m} u_m[-\text{log}(CX (\phi_m(\hat{x}_B), \phi_m (y_B))) \nonumber \\
	& + \sum\limits_m \eta_m \| \phi_m(\hat{x}_B) - \phi_m (\tilde{y}_B)\|_1,
\end{align}
where $u_m$ controls the relative importance of different VGG layers
and $\eta_m$ is the  balancing coefficient.

\noindent {\bf Adversarial loss.}
We add a discriminator to distinguish outputs from the real images in the target domain, competing with the generator which tries to synthesize images that are indistinguishable. The adversarial loss is,
\begin{align}
	\mathcal{L}_{adv}^{\mathcal{D}} & = - \mathbb{E}[h(\mathcal{D}(y_B))] 
	- \mathbb{E}[h(-\mathcal{D}(\mathcal{G}(x_A,y_B)))], \\
	\mathcal{L}_{adv}^{\mathcal{G}} & = - \mathbb{E}[\mathcal{D}(\mathcal{G}(x_A,y_B))],
\end{align}
where $h(t)=\min(0,-1+t)$ is the hinge loss~\cite{zhang2019self,brock2018large} to regularize the discriminator.

\noindent {\bf Total loss.}
In summary, our overall objective function is,
\begin{align}
\mathcal{L} &= \min_{\mathcal{M},\mathcal{N},\mathcal{G}, \mathcal{R}} \max_{\mathcal{D}}\lambda_1\mathcal{L}_{align}+\lambda_2 \mathcal{L}_{corr} + \lambda_3\mathcal{L}_{map}
\nonumber \\
&+ \lambda_4 (\mathcal{L}_{sem} + \mathcal{L}_{app}) 
+\lambda_5 (\mathcal{L}_{adv}^{\mathcal{D}}+\mathcal{L}_{adv}^{\mathcal{G}}),
\end{align}
where $\lambda$ denotes the weighting parameters, $\mathcal{M}$ contains $\mathcal{M}_A$ and $\mathcal{M}_B$, and $\mathcal{N}$ includes $\mathcal{N}_1, \cdots, \mathcal{N}_{L}$.

\begin{figure*}[t]
\centering
\small
\begin{tabularx}{0.85\textwidth}{YYYYYY}
Input & Ground truth & SPADE & CoCosNet & \emph{CoCosNet v2} & Exemplar
\end{tabularx}
\renewcommand{\arraystretch}{0.0}
\begin{tabular}{c}
\includegraphics[width=1.8\columnwidth]{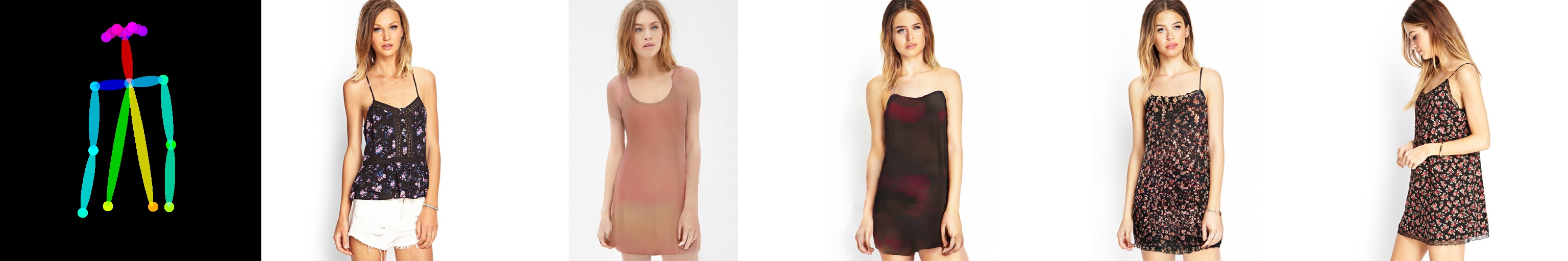}\\
\includegraphics[width=1.8\columnwidth]{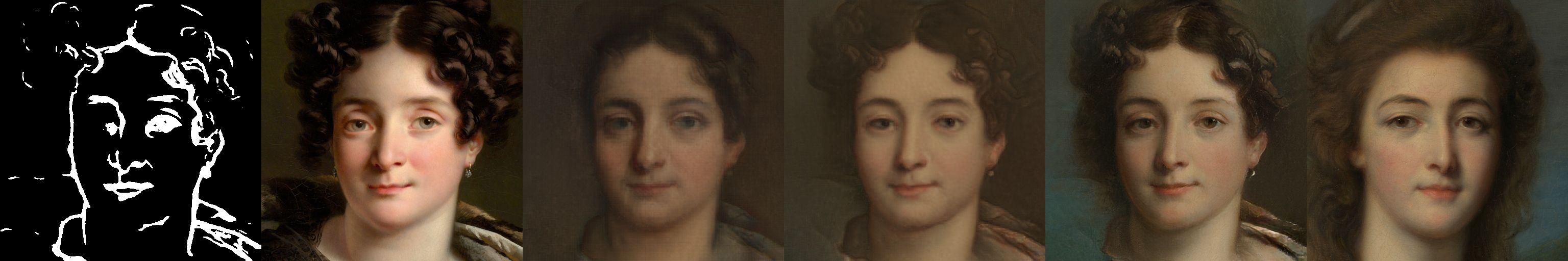}\\
\includegraphics[width=1.8\columnwidth]{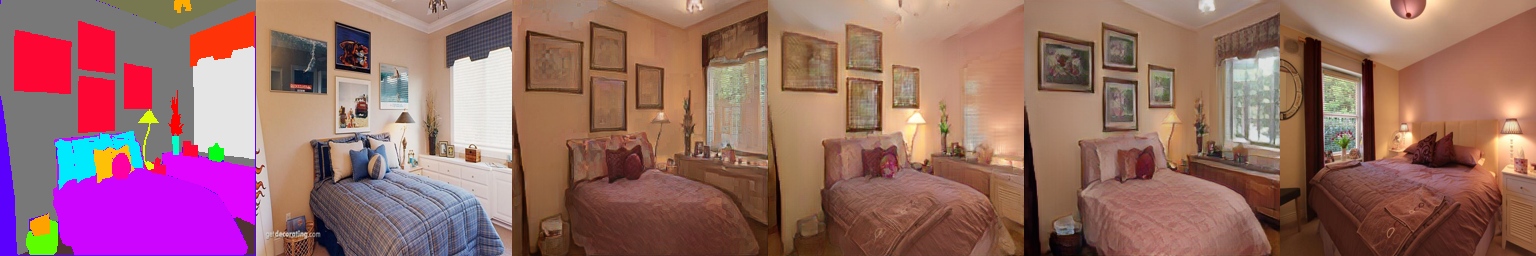}
\end{tabular}
\caption{{Qualitative comparison on the Deepfashion dataset, the MetFaces dataset, and the ADE20K dataset respectively.}}
\vspace{-5mm}
\label{figure:qualitative_comparison}
\end{figure*}

%------------------------------------------------------------------------

\section{Experiment}

\noindent {\bf Implementation details.}
We apply spectral normalization~\cite{miyato2018spectral} to all the layers for the translation network and discriminator. We use the Adam solver~\cite{kingma2014adam} with $\beta_1=0$ and $\beta_2=0.999$.
The learning rates for the generator and the discriminator are set as $1e-4$ and $4e-4$ respectively, following the TTUR~\cite{heusel2017gans}.
For detailed implementation including network architectures, please see our appendix.
The experiments are conducted using $8$ $32$GB Tesla V100 GPUs.

\noindent {\bf Datasets.}
We conduct experiments on four datasets: 
\begin{itemize}[leftmargin=*,topsep=1pt]
\itemsep-0.1em
\item {DeepFashion}~\cite{liu2016deepfashion} consists of $52,712$ high-quality fashionable person images. 
We adopt the high-resolution version, and conduct pose-to-body synthesis at 512$\times$512 resolution. 
OpenPose~\cite{cao2017realtime} is used for pose extraction.
\item {MetFaces}~\cite{karras2020training} consists of $1,336$ high-quality human face images at $1024\times 1024$ resolution collected from works of art in the Metropolitan Museum. The images in the dataset exhibit a wide variety in artistic style. 
We use the HED~\cite{xie2015holistically} to obtain the background edges and connect the face landmarks for the face region. On this dataset, we learn the translation from edges to faces.
\item {ADE20K}~\cite{zhou2017scene} consists of $20,210$ training and $2,000$ validation images. Each image is paired with a 150-class segmentation mask.
Because of its large diversity, it is challenging for most existing methods to perform mask-to-scene translation.
As most of the images have short side $<$512, we synthesize images at resolution 256$\times$256 on this dataset.
\item {ADE20K-outdoor} is the subset of ADE$20K$.
We follow the same protocol in SIMS~\cite{qi2018semi}.
\end{itemize}

\begin{table}[!t]
\footnotesize
\centering
\setlength\tabcolsep{3.0pt}
\begin{tabular}{@{}lcccccccc@{}}
\toprule
\multirow{2}{*}{} & \multicolumn{2}{c}{DeepFashion} & \multicolumn{2}{c}{MetFaces} & \multicolumn{2}{c}{ADE20k} &\multicolumn{2}{c}{ADE20k-\emph{outdoor}}\\
\cmidrule(lr){2-3}
\cmidrule(lr){4-5}
\cmidrule(lr){6-7}
\cmidrule(lr){8-9}
&FID    &SWD    &FID    &SWD    &FID    &SWD    &FID    &SWD\\
\midrule
SPADE    &34.4    &38.0    &39.8    &30.4    &33.9 &19.7 &63.3 &21.9\\
CocosNet    &26.9    &29.0    &25.6    &24.3    &26.4 &10.5 &42.4 &11.5\\
\emph{CoCosNet v2}    &\textbf{22.5}    &\textbf{24.6}    &\textbf{23.3}    &\textbf{22.4}    &\textbf{25.2}    &\textbf{9.9}    &\textbf{38.9}    &\textbf{10.2}\\
\bottomrule
\end{tabular}
\caption{Quantitative evaluation of image quality. For both metrics, the lower is better, with the best scores highlighted. }
\vspace{-3mm}
\label{table:quantitative_comparison}
\end{table}

\begin{table}[!t]
\footnotesize
\setlength\tabcolsep{5.5pt}
\centering 
\begin{tabular}{@{}lcccc@{}}
\toprule
& DeepFashion & MetFaces & ADE20k & ADE20k-\emph{outdoor} \\
\midrule
SPADE     & 0.883 & 0.915 &0.856 &0.867\\
CoCosNet     & 0.924 & 0.941 &0.862 &0.873\\
\emph{CoCosNet v2} & \textbf{0.959} &\textbf{0.963} & \textbf{0.877} & \textbf{0.895}\\
\bottomrule
\end{tabular}
\caption{Quantitative evaluation of semantic consistency. The higher is better with the best scores highlighted.
} 
\vspace{-3mm}
\label{table:semantic_consistency}
\end{table}

\begin{table}[!t]
\footnotesize
\centering
\setlength\tabcolsep{6.4pt}
\begin{tabular}{@{}lcccccc@{}}
\toprule
\multirow{2}{*}{} & \multicolumn{2}{c}{DeepFashion} & \multicolumn{2}{c}{MetFaces} & \multicolumn{2}{c}{ADE20k}\\
\cmidrule(lr){2-3}
\cmidrule(lr){4-5}
\cmidrule(lr){6-7}
& Color & Texture & Color & Texture & Color & Texture\\
\midrule
SPADE       &0.932 &0.893 &0.949 &0.920 &0.874 &0.892\\
CoCosNet    &0.975 &0.944 &0.956 &0.932 &0.962 &0.941\\
\emph{CoCosNet v2} &\textbf{0.987} &\textbf{0.961} & \textbf{0.972} & \textbf{0.956} & \textbf{0.970} & \textbf{0.948}\\
\bottomrule
\end{tabular}
\caption{Quantitative evaluation of style relevance. The higher is better with the best scores highlighted.
}
\vspace{-5mm}
\label{table:style_relevance}
\end{table}

\begin{figure*}[t]
\centering
\renewcommand{\arraystretch}{0.0}
\begin{tabular}{c}
\includegraphics[width=2.0\columnwidth]{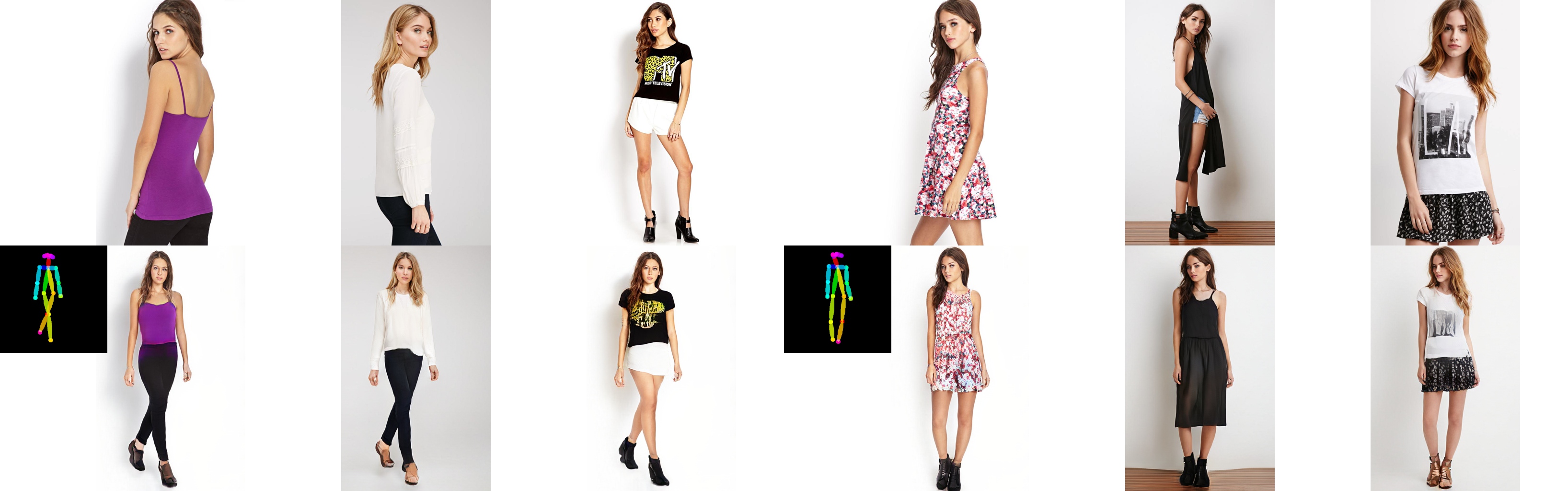}\\
\includegraphics[width=2.0\columnwidth]{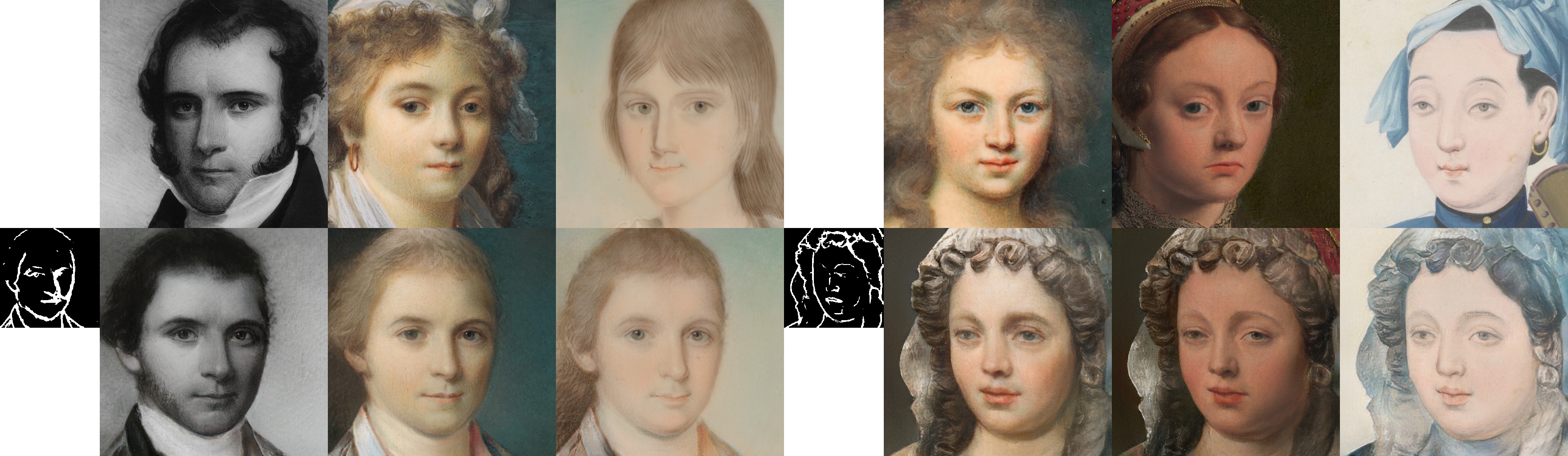}
\end{tabular}
\caption{More results at resolution of 512$\times$512 by CoCosNet v2. For each group, 1st row: exemplars; 2nd row: our results.}
\vspace{-5mm}
\label{figure:more_results1}
\end{figure*}

\begin{figure}[t]
\centering
\renewcommand{\arraystretch}{0.0}
\begin{tabular}{c}
\includegraphics[width=0.8\columnwidth]{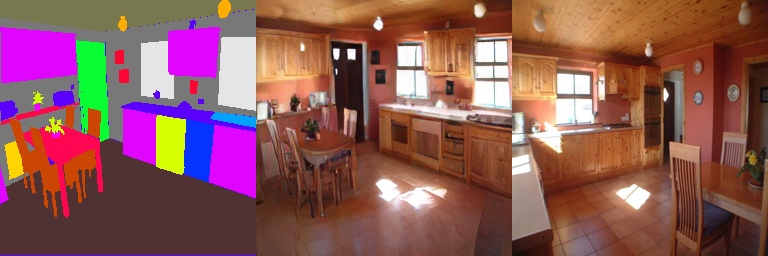}\\
\includegraphics[width=0.8\columnwidth]{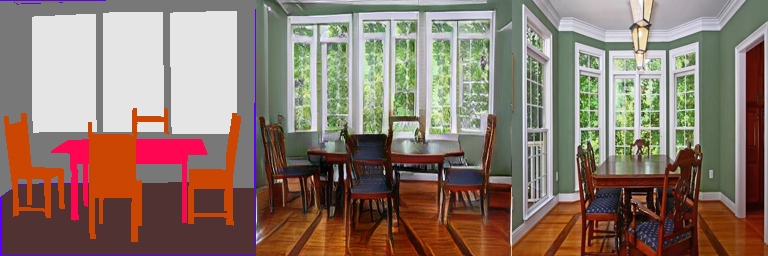}\\
\includegraphics[width=0.8\columnwidth]{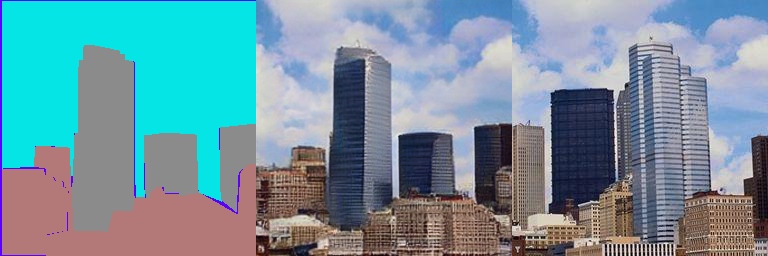}\\
\end{tabular}
\caption{Our results on the ADE20k dataset. Left to right: input, our results, the exemplar.}
\vspace{-6mm}
\label{figure:more_results}
\end{figure}

%-------------------------------------------------------------------------
\subsection{Comparison with the state-of-the-Art}
There are many excellent works that have been proposed for general image translation.
We do not compare with those methods that
directly learn the translation through networks and fail to utilize the style of exemplars, such as Pix2pixHD~\cite{wang2018high}
and SIMS~\cite{qi2018semi}.
We compare with two strong baselines. One is the SPADE~\cite{park2019semantic}, a leading approach among the methods~\cite{ma2018exemplar,huang2017arbitrary,huang2018multimodal} that leverage the exemplar style in a global way. 
We also compare our method with the closest competitor CoCosNet~\cite{zhang2020cross} that also leverages cross-domain correspondence but learns at low-resolution. 
The two works are initially proposed for generating images at resolution $256\times 256$. For a fair comparison, we retrain their models on Deepfashion and MetFaces at resolution $512\times 512$ and make appropriate modifications in order to generate high-quality translation results.

\begin{table}[!t]
\footnotesize
\centering 
\setlength\tabcolsep{6.0pt}
\begin{tabular}{@{}lccc@{}}
\toprule
& L1$\downarrow$ &PSNR$\uparrow$ &SSIM$\uparrow$ \\
\midrule
64 &82.25 &28.03 &0.75 \\
64+128  &79.56 &28.09 &0.76 \\
64+128+256 &79.10 &29.50 &0.79 \\
\emph{Full} 64+128+256+512  &\textbf{77.84} &\textbf{30.03} &\textbf{0.82} \\
\bottomrule
\end{tabular}
\caption{Ablation study of full-resolution correspondence.}
\vspace{-3mm}
\label{table:ablation_resolution}
\end{table}

\begin{table}[!t]
\footnotesize
\centering 
\setlength\tabcolsep{6.0pt}
\begin{tabular}{@{}lccc@{}}
\toprule
& L1$\downarrow$ &PSNR$\uparrow$ &SSIM$\uparrow$ \\
\midrule
Only PatchMatch propagation&108.75 &20.40 &0.67 \\
Only ConvGRU  &94.21 &22.99 &0.74 \\
PatchMatch propagation + conv  &87.83 &23.54 &0.76 \\
PatchMatch propagation + ConvGRU (\emph{ours})  &\textbf{81.97} &\textbf{28.99} &\textbf{0.83} \\
\bottomrule
\end{tabular}
\caption{Ablation study of GRU-assisted refinement.}
\vspace{-5mm}
\label{table:ablation_gru}
\end{table}

\begin{figure*}[t]
\centering
\includegraphics[width=2\columnwidth]{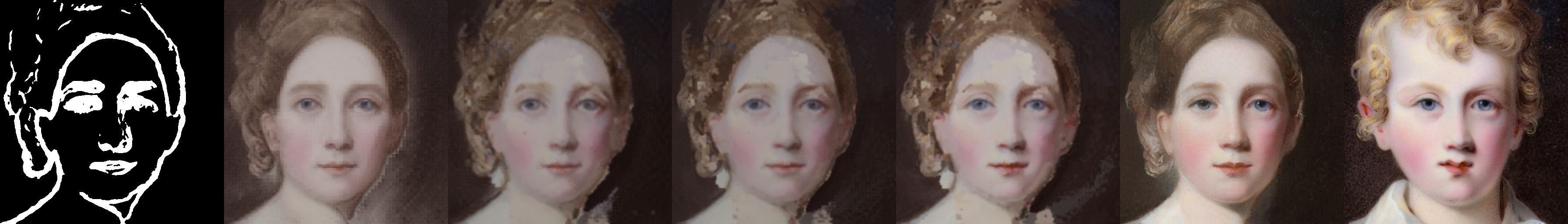}
\caption{Comparison of warped images at different resolution levels. From left to right: edge, warped images at $64^2$, $128^2$, $256^2$, $512^2$, output, exemplar. The warped image at $512^2$ exhibits more details.}
\label{figure:ablation_hierarchical}
\end{figure*}

\begin{figure*}[t]
\centering
\includegraphics[width=2\columnwidth]{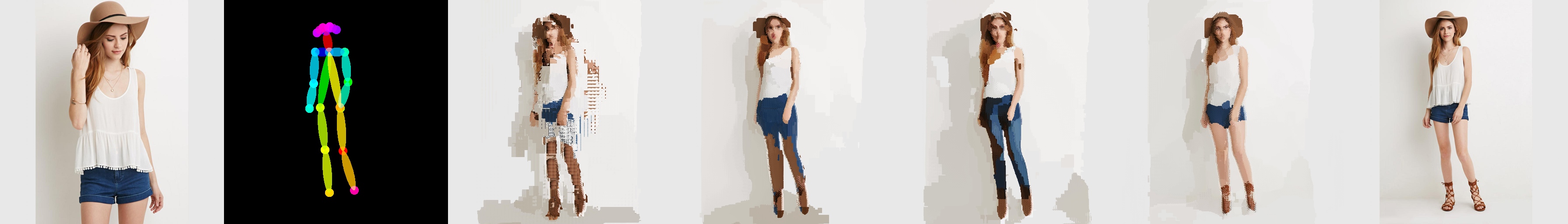}
\caption{Comparison of warped images for different variants of GRU-assisted refinement. From left to right: exemplar, pose, warped images for using only PatchMatch propagation, only ConvGRU, PatchMatch propagation with convolution, CoCosNet v2 using PatchMatch propagation with convGRU, and ground truth. CoCosNet v2 produces the most faithful warping image.}
\vspace{-1.5em}
\label{figure:gru}
\end{figure*}

\begin{figure}[t]
\centering
\includegraphics[width=0.95\columnwidth]{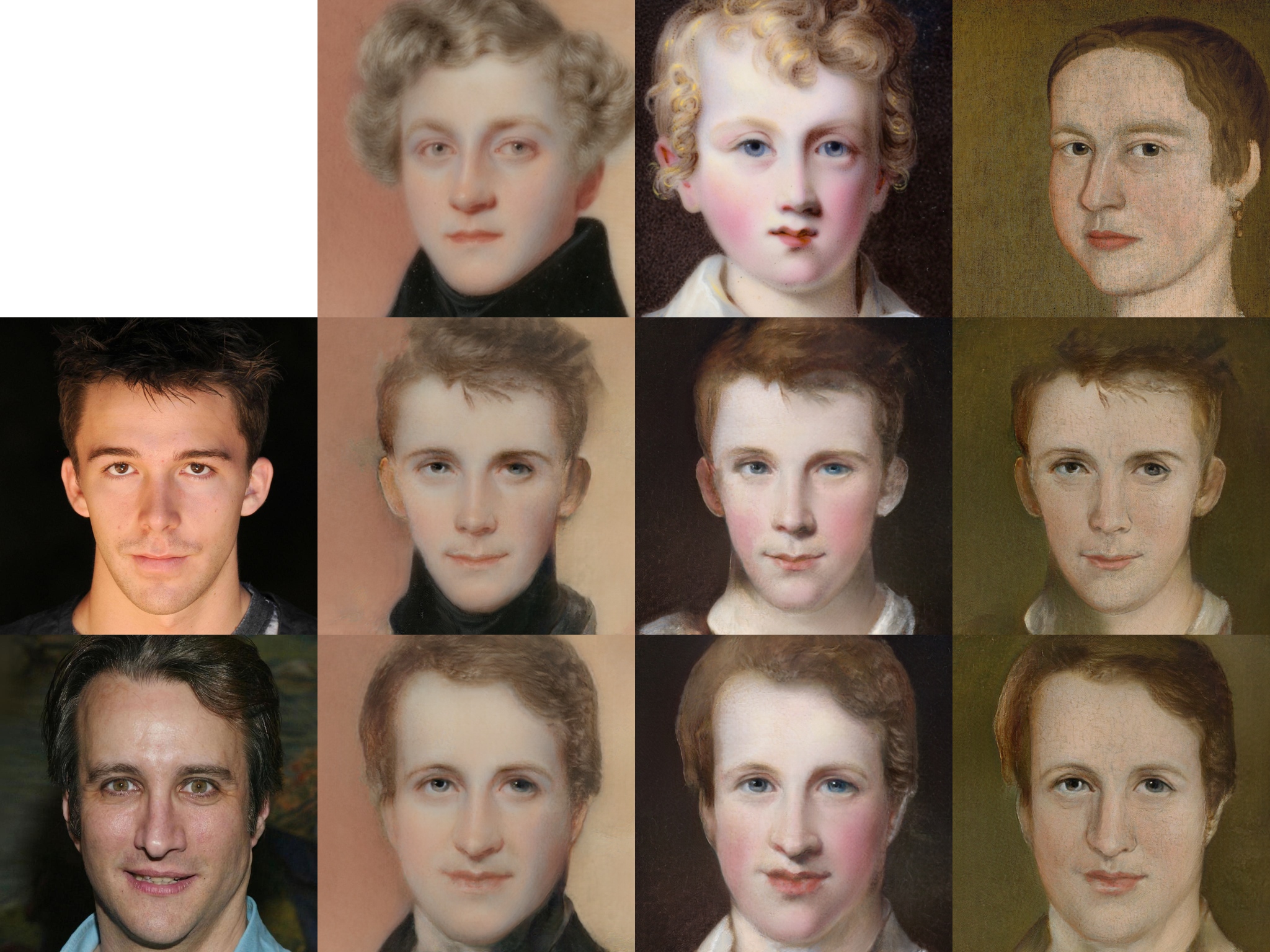}
\caption{Oil portrait. Given a portrait, CoCosNet v2 can transfer it to a customized oil painting with style from a given exemplar.}
\label{figure:oil_portrait}
\vspace{-2em}
\end{figure}

\noindent {\bf Quantitative evaluation.}
We first present quantitative evaluation from three directions following~\cite{zhang2020cross}.
(1) Image quality is evaluated with
two widely adopted metrics. One is Fr$\acute{\text{e}}$chet Inception Distance score (FID)~\cite{heusel2017gans} that aims to calculate the distance between Gaussian fitted feature distributions of real and generated images. The other one is sliced Wasserstein distance (SWD)~\cite{karras2017progressive} that attempts to measure the Wasserstein distance between the distributions of real images and synthesized ones.
Both metrics have been shown that a lower score indicates higher quality images;
(2) Semantic consistency is evaluated between the output and the input by calculating the cosine similarity between high-level features representing semantics, \ie,  $relu3\_2$, $relu4\_2$ and $relu5\_2$ of an ImageNet pretrained VGG model~\cite{brock2018large};
(3) Style relevance is evaluated between the output and the exemplar with low-level features, $relu1\_2$, and $relu2\_2$ that mostly encode the color and texture information.
The comparison results are shown in Table~\ref{table:quantitative_comparison}, Table~\ref{table:semantic_consistency}, and Table~\ref{table:style_relevance} respectively. 
We can see that CoCosNet v2 significantly outperforms prior competitive methods in the three aspects, suggesting that CoCosNet v2 synthesizes images of higher quality, better preserved semantics and more relevant style.

\noindent {\bf Qualitative comparison.}
We show qualitative comparison with the competitors in Figure~\ref{figure:qualitative_comparison}. It can be clearly seen that CoCosNet v2 produces the most visually appealing results and the least visible artifacts. 
We find that the distinctive patterns in the exemplar have been remarkably well preserved in the semantically corresponding regions of the output, 
\eg, the texture patterns of the dress in pose-to-body translation, which has been washed out in SPADE and CoCosNet.
On the other hand, our output depicts subtle details that are of particular importance to a high-resolution image, demonstrating the advantage of CoCosNet v2.
Figure~\ref{figure:more_results1}-\ref{figure:more_results} shows more diverse results under different exemplars. We also demonstrate 1024$\times$1024 results in Figure~\ref{figure:teaser}.

\subsection{Ablation study}

\noindent {\bf Full-resolution correspondence.}
We validate the effectiveness of full-resolution correspondence, which benefits CoCosNet v2 in producing fine textures in the ultimate output. We explore the translation results when correspondence is established at certain level of limited resolution.
Specifically, we vary the largest resolution, \ie the dimension of the latent space, from $64^2$ to $512^2$ and see how the performance changes. We evaluate the warping on Deepfashion dataset as we consider the person image under a different pose as the exemplar as well as the ground truth. Hence, we can measure the warping with  L1, PSNR and SSIM~\cite{wang2004image}. The numerical results in Table~\ref{table:ablation_resolution} show that hierarchical PatchMatch offers a more accurate correspondence in high-resolution. The qualitative study in Figure~\ref{figure:ablation_hierarchical} shows that full-resolution correspondence captures more details,
which further benefits the high-quality synthesis.

\noindent {\bf GRU-assisted refinement.}
We present a comprehensive analysis to justify the important component in our architecture, \ie GRU-assisted refinement.
We study three variants that are different in each iteration: 1) using only PatchMatch propagation;
2) using only ConvGRU refinement; 3) using PatchMatch propagation assisted with convolution.
The comparison with our full model (PatchMatch propagation assisted with ConvGRU) are presented
in Table~\ref{table:ablation_gru} numerically and Figure~\ref{figure:gru} visually.
We can see that only adopting PatchMatch propagation or ConvGRU produces inferior results. We conjecture that the reason may be 1) only PatchMatch propagation cannot backward the gradient to the correctly matched patches, and hence get trapped in the local optimum; 2) only ConvGRU does not consider neighborhood coherence and thus fails to preserve local textures.
Moreover, we find that CoCosNet v2 is better than the third variant, which demonstrates that ConvGRU plays an important role in utilizing the history information.
\vspace{-2mm}

\subsection{Application of oil portrait}
\vspace{-2mm}

We present an intriguing application of oil portrait that transfers a portrait to a custom oil painting with different styles specified by the exemplar. This is achieved by extracting the edges from real faces, \eg, images from CelebA~\cite{liu2015deep}, and applying the model trained from MetFaces. We show several examples in Figure~\ref{figure:oil_portrait}.
\vspace{-2mm}

%-------------------------------------------------------------------------
\section{Conclusion}
\vspace{-2mm}

We propose to learn the semantic correspondence in full-resolution.
To achieve that, we introduce an effective algorithm CoCosNet v2 that efficiently establishes the correspondence through iterative refinement in a coarse-to-fine hierarchy. 
At each level, the propagation and GRU-based propagation are alternatively performed. 
CoCosNet v2 leads to photo-realistic outputs with fine textures as well as visually appealing images at large resolutions, $512^2$ and $1024^2$.

%-------------------------------------------------------------------------

\small{
\noindent {\bf Acknowledgments. }This work is supported in part by Science and Technology Innovation 2030 – ``New Generation Artificial Intelligence'' Major Project No.(2018AAA0100904), NSFC (No. U19B2043), Artificial Intelligence Research Foundation of Baidu Inc., the funding from HIKVision and Horizon Robotics.
}

%-------------------------------------------------------------------------
% SUPP
%-------------------------------------------------------------------------
% \cleardoublepage
\begin{onecolumn}

\label{sec:supp}
\setcounter{section}{0}
\renewcommand\thesection{\Alph{section}}
\renewcommand \thesubsection {\arabic{subsection}}
\vspace{-4em}

% {\Large Full-Resolution Correspondence Learning for Image Translation}

\noindent{\large{\textbf{Appendix of Full-Resolution Correspondence Learning for Image Translation}}}

%------------------------------------------------------------------------
\section{Additional Generation Results}
\subsection{Pose-to-body}

Figure~\ref{figure:pose_to_image_01} to Figure~\ref{figure:pose_to_image_05} show more results about pose-to-body generation at the resolution $512\times512$ on the Deepfashion dataset. To the best of our knowledge, our approach is the first work to generate person images at the resolution $512\times512$ on the Deepfashion dataset. Our approach is able to well preserve the patterns, \ie, logos and letters, on the clothing because of the \emph{full-resolution} correspondences constructed between two images. The person images generated by our approach are highly authentic and vivid.

\begin{figure*}[hbtp]
\centering
\includegraphics[width=0.9\columnwidth]{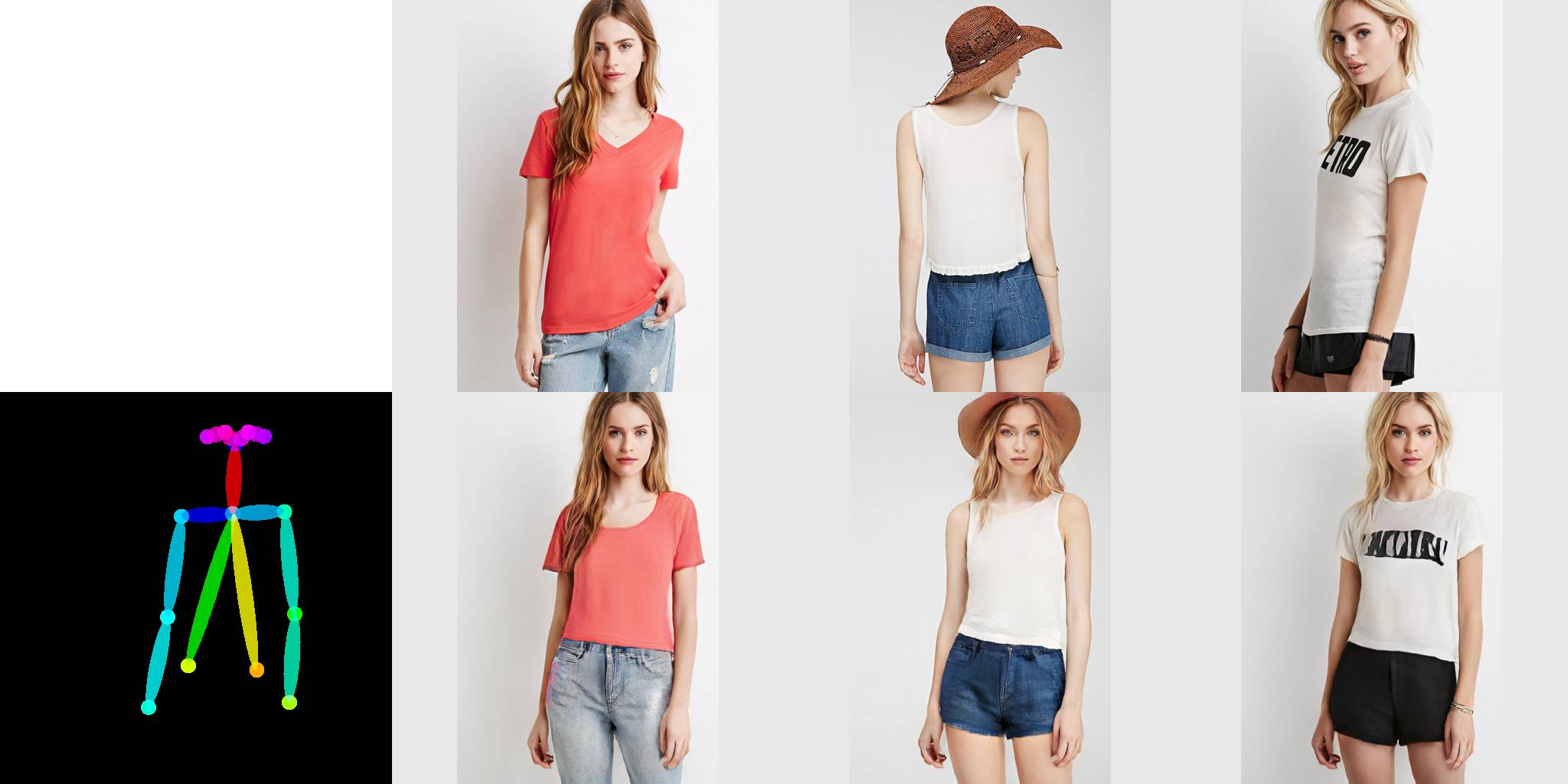}\\
\vspace{+1em}
\includegraphics[width=0.9\columnwidth]{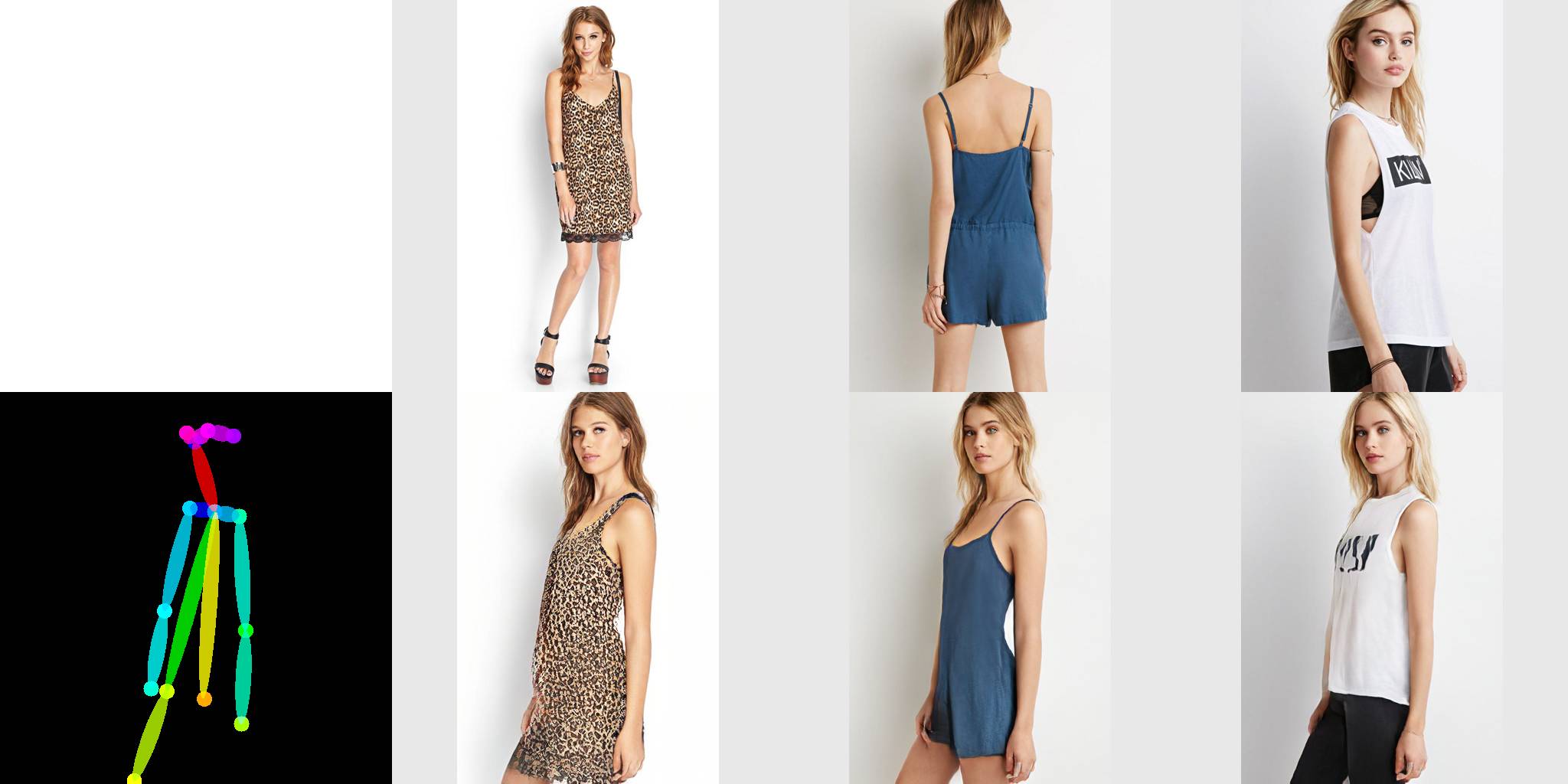}\\
\vspace{+1em}
\caption{Pose-to-body image translation results at resolution $512\times512$. 1st row: exemplar images, 2nd row: generated images.  (Deepfashion dataset)}
\label{figure:pose_to_image_01}
\end{figure*}

\clearpage
\begin{figure*}
\centering
\includegraphics[width=0.9\columnwidth]{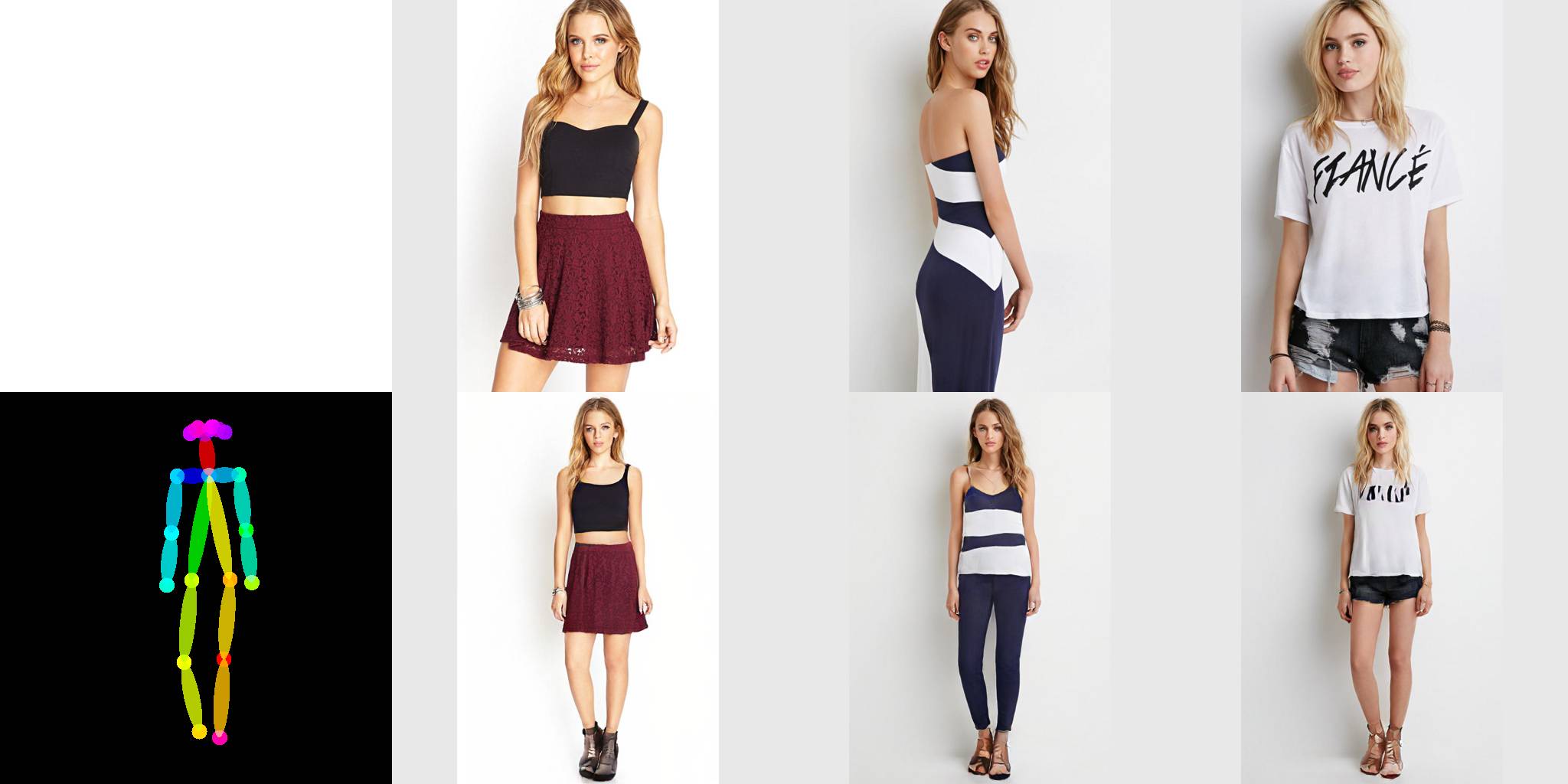}\\
\vspace{+3em}
\includegraphics[width=0.9\columnwidth]{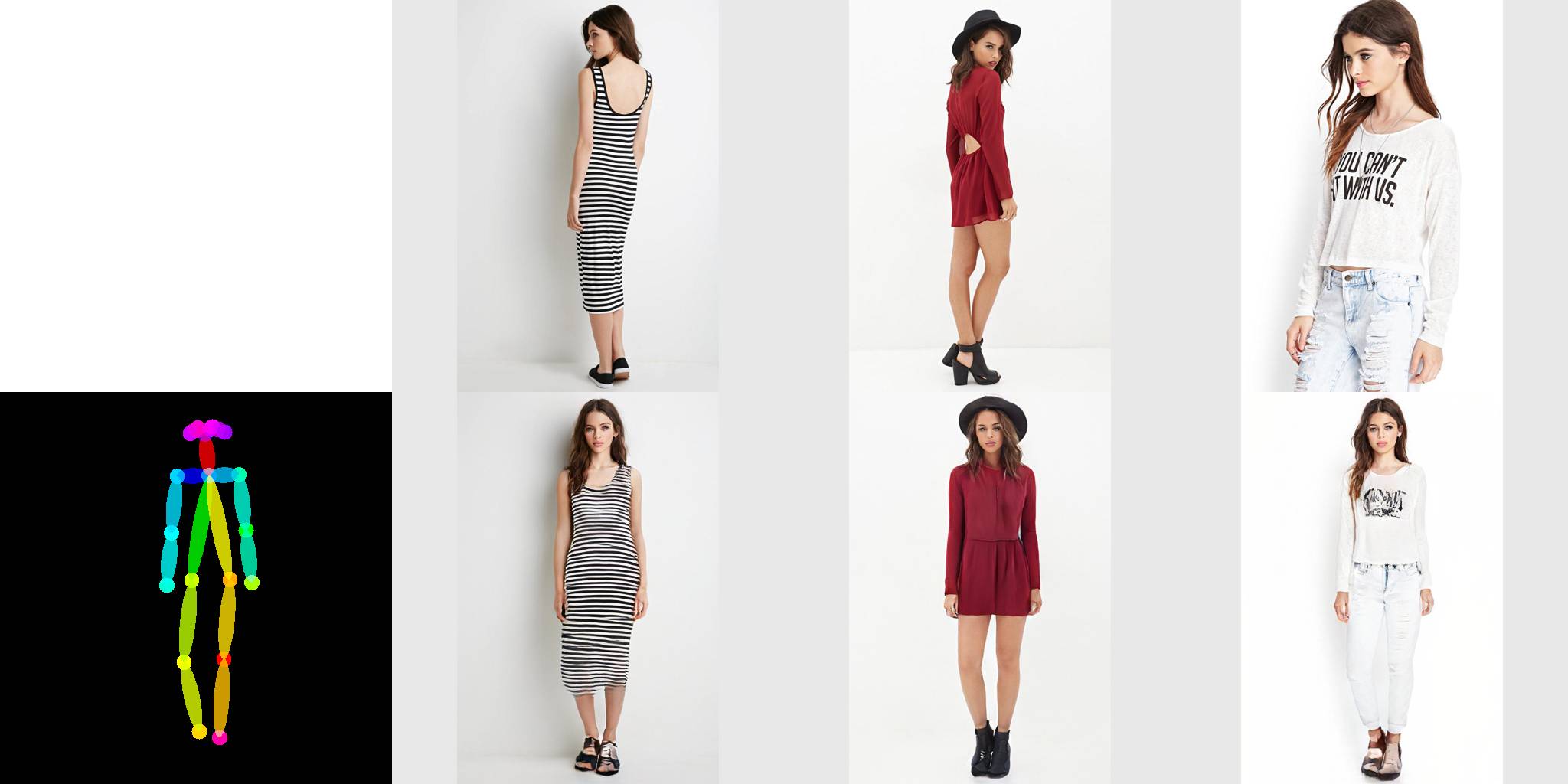}\\
\vspace{+1em}
\caption{Pose-to-body image translation results at resolution $512\times512$. 1st row: exemplar images, 2nd row: generated images.  (Deepfashion dataset)}
\label{figure:pose_to_image_02}
\end{figure*}

\clearpage
\begin{figure*}
\centering
\includegraphics[width=0.9\columnwidth]{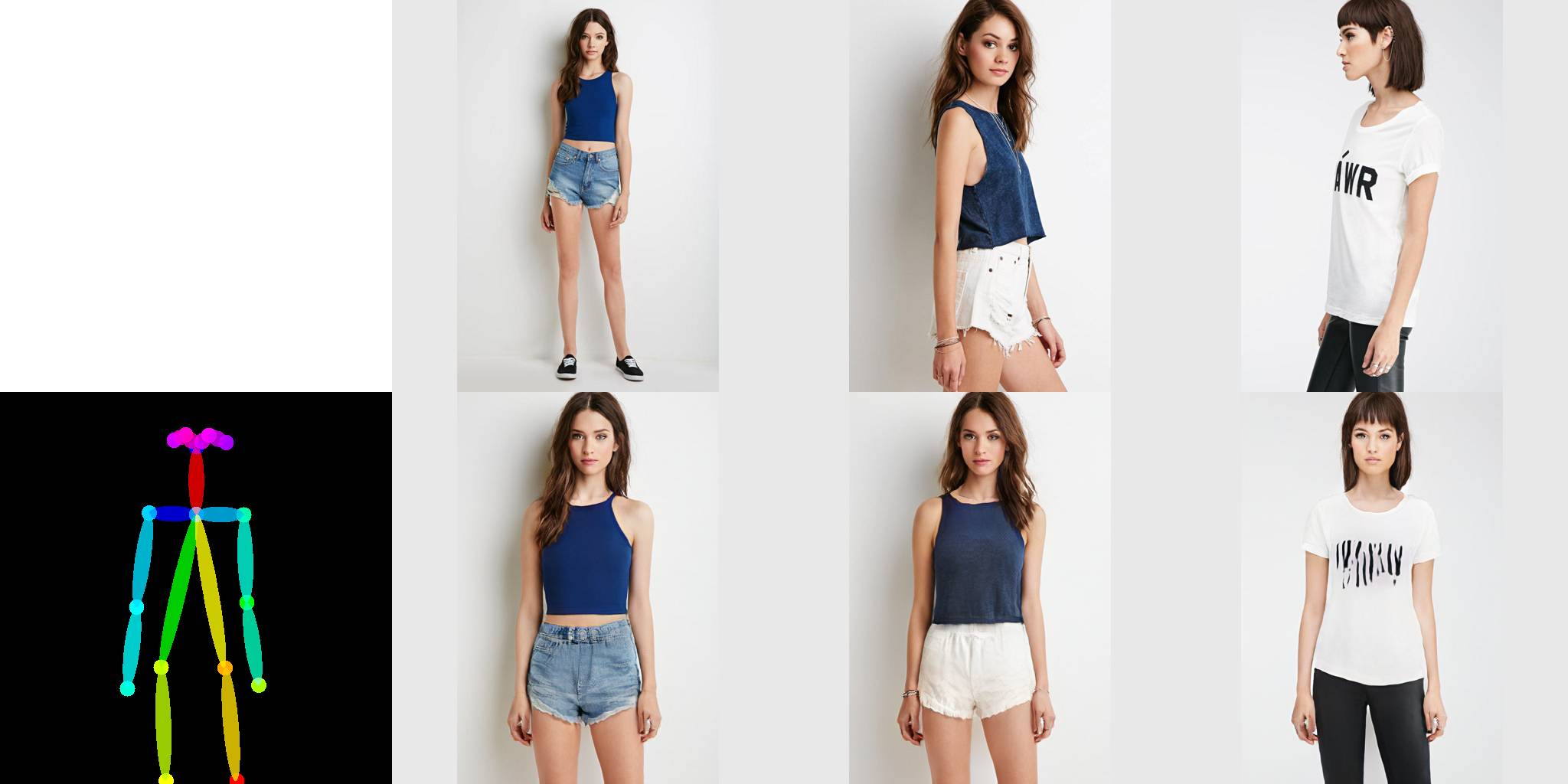}\\
\vspace{+3em}
\includegraphics[width=0.9\columnwidth]{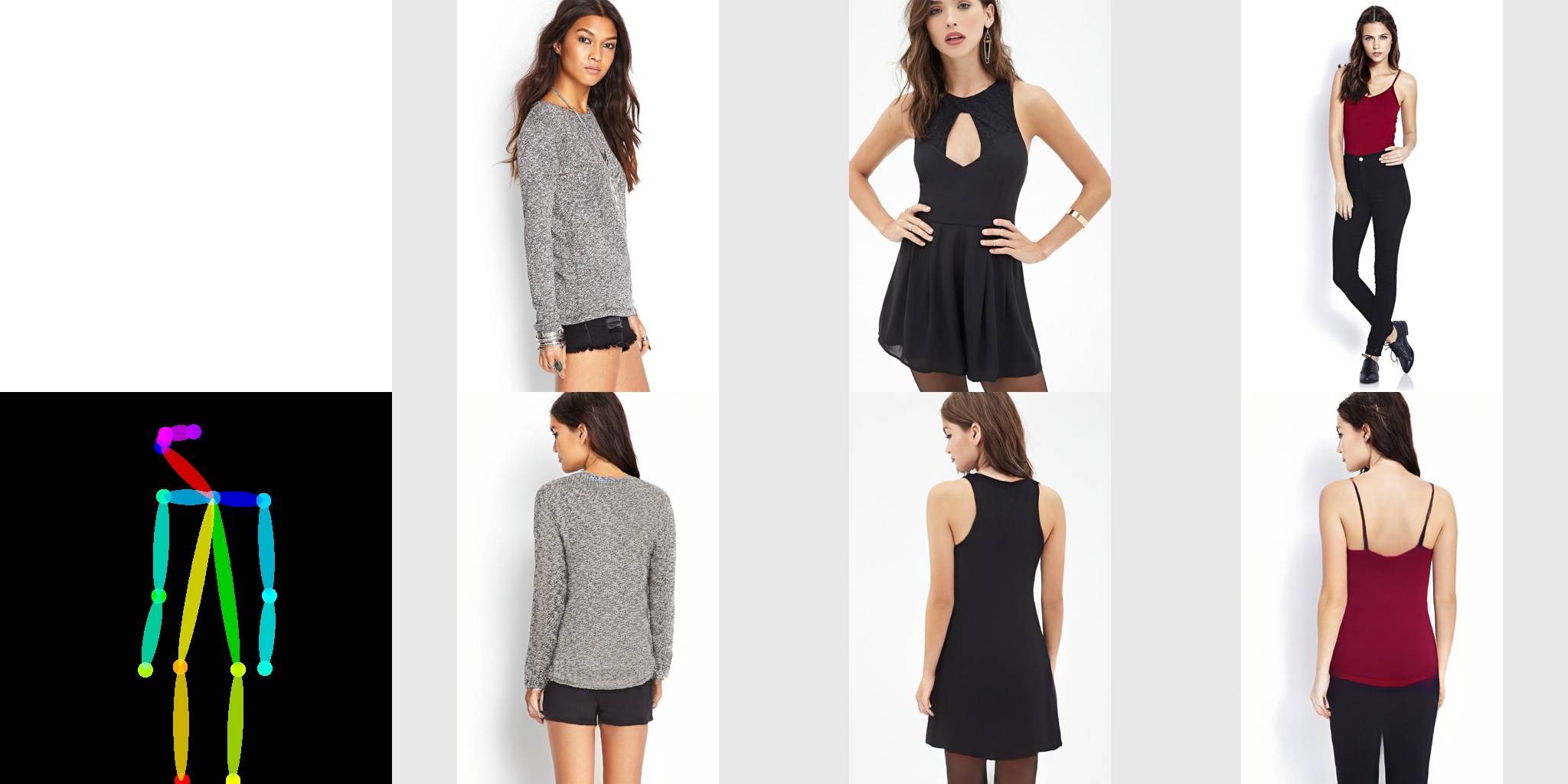}\\
\vspace{+1em}
\caption{Pose-to-body image translation results at resolution $512\times512$. 1st row: exemplar images, 2nd row: generated images.  (Deepfashion dataset)}
\label{figure:pose_to_image_03}
\end{figure*}

\clearpage
\begin{figure*}
\centering
\includegraphics[width=0.9\columnwidth]{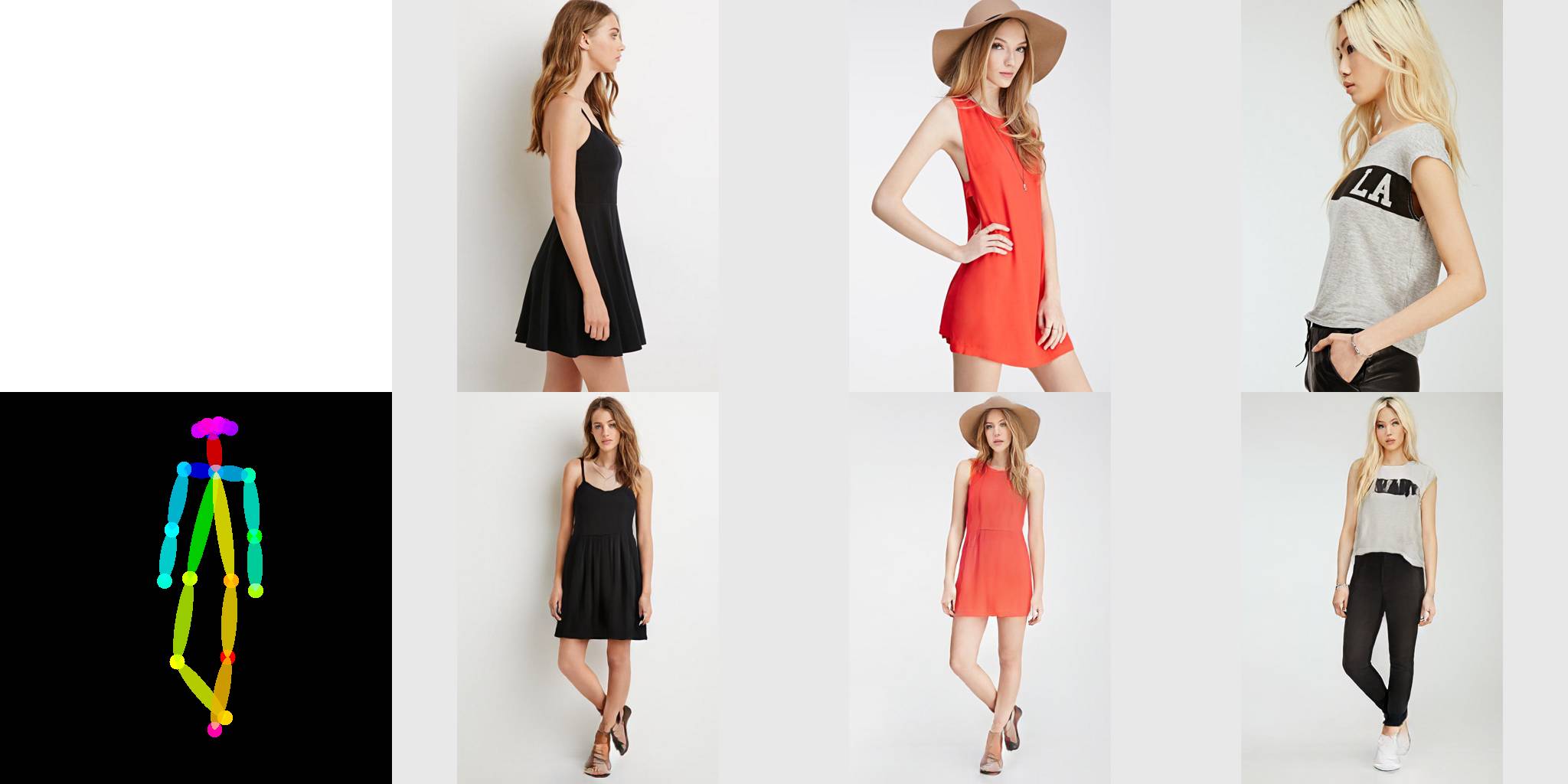}\\
\vspace{+3em}
\includegraphics[width=0.9\columnwidth]{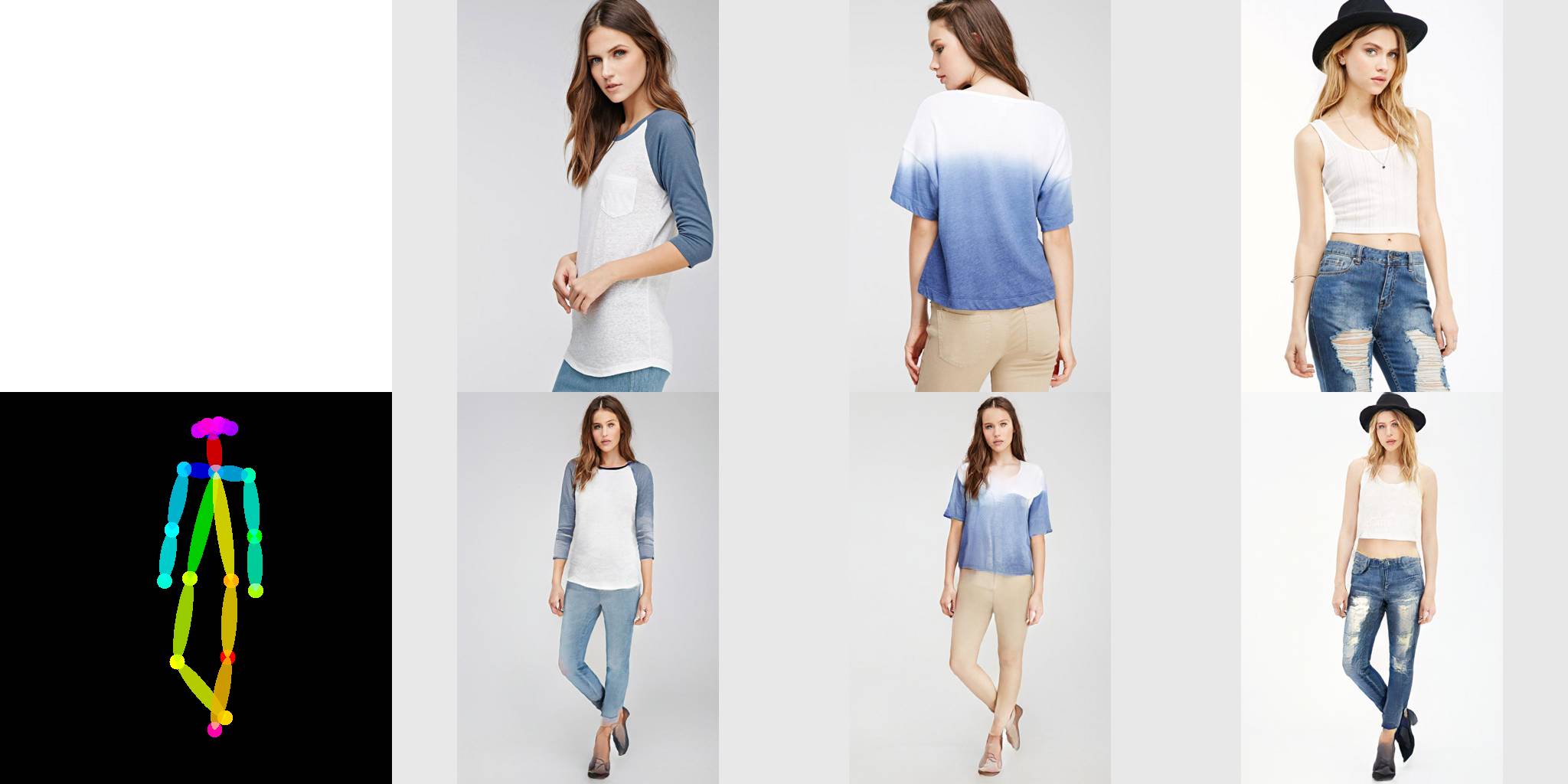}\\
\vspace{+1em}
\caption{Pose-to-body image translation results at resolution $512\times512$. 1st row: exemplar images, 2nd row: generated images.  (Deepfashion dataset)}
\label{figure:pose_to_image_04}
\end{figure*}

\clearpage
\begin{figure*}
\centering
\includegraphics[width=0.9\columnwidth]{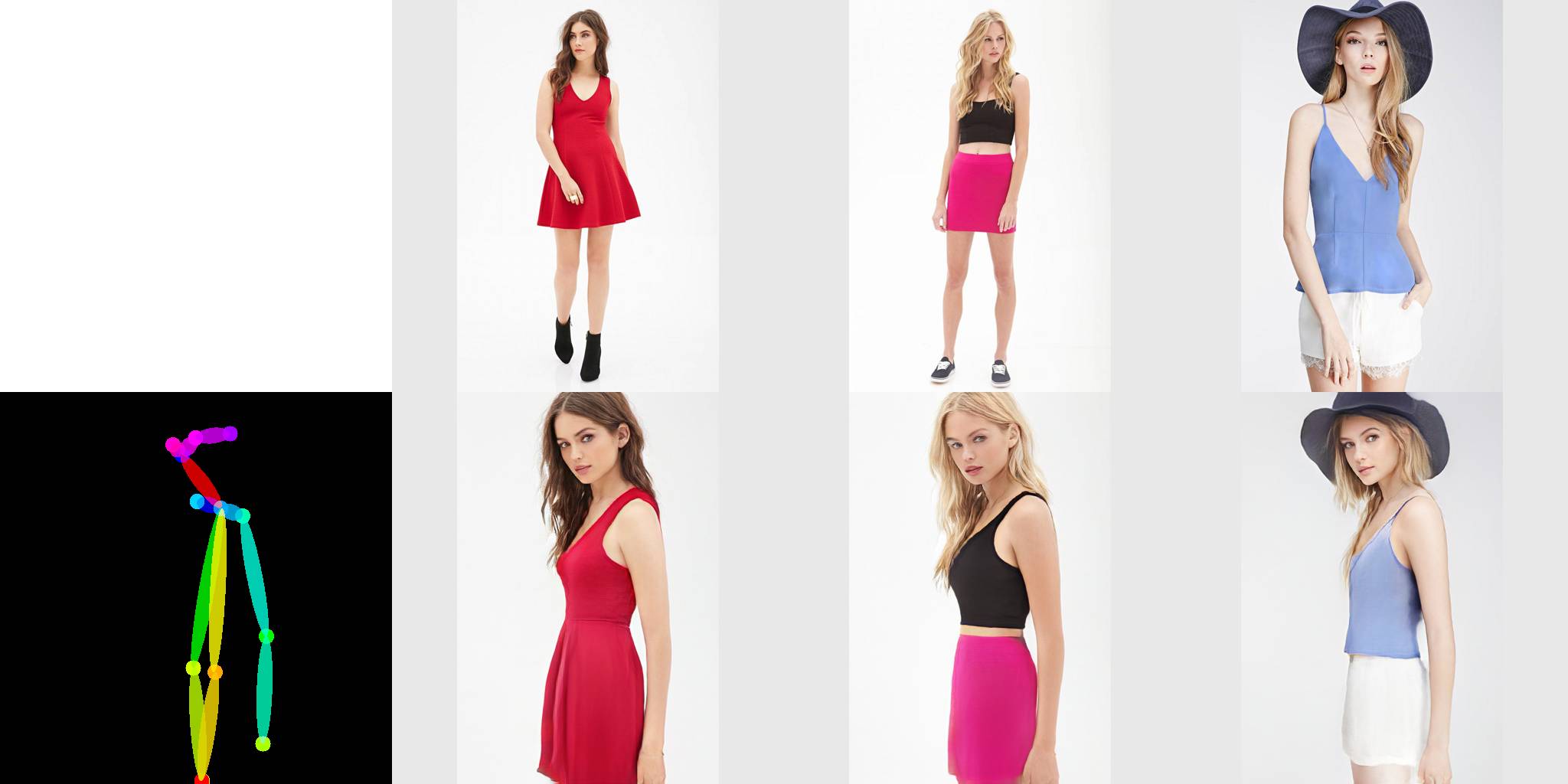}\\
\vspace{+3em}
\includegraphics[width=0.9\columnwidth]{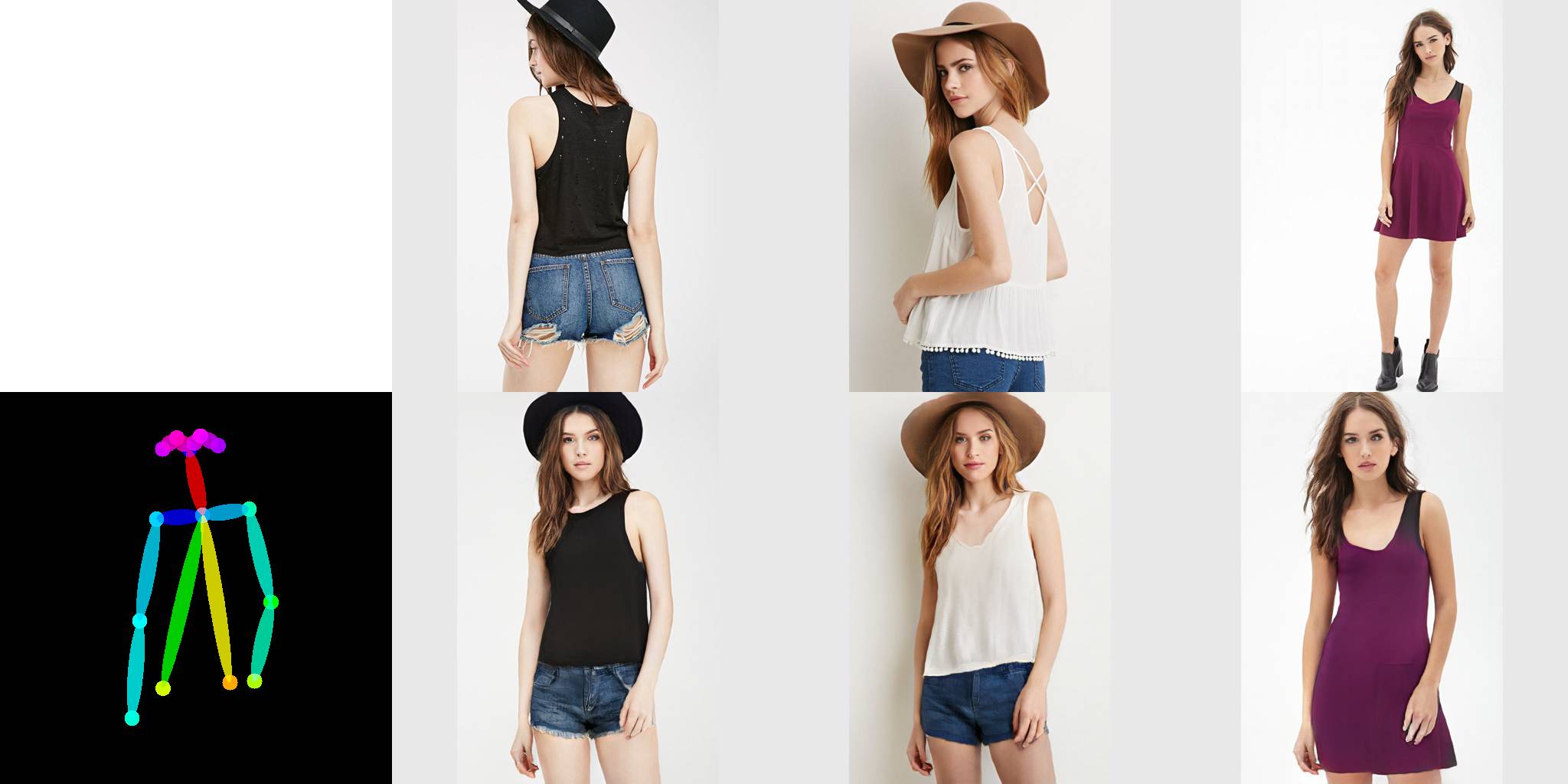}\\
\vspace{+1em}
\caption{Pose-to-body image translation results at resolution $512\times512$. 1st row: exemplar images, 2nd row: generated images.  (Deepfashion dataset)}
\label{figure:pose_to_image_05}
\end{figure*}

\clearpage
\subsection{Edge-to-face}

Figure~\ref{figure:edge_to_face_01} to Figure~\ref{figure:edge_to_face_04} show more results of edge-to-face generation at the resolution $1024\times1024$ on the MetFaces dataset. Our approach produces visually appealing edge-to-face translation results at high-resolution.

\vspace{+5em}
\begin{figure*}[hbtp]
\centering
\includegraphics[width=1\columnwidth]{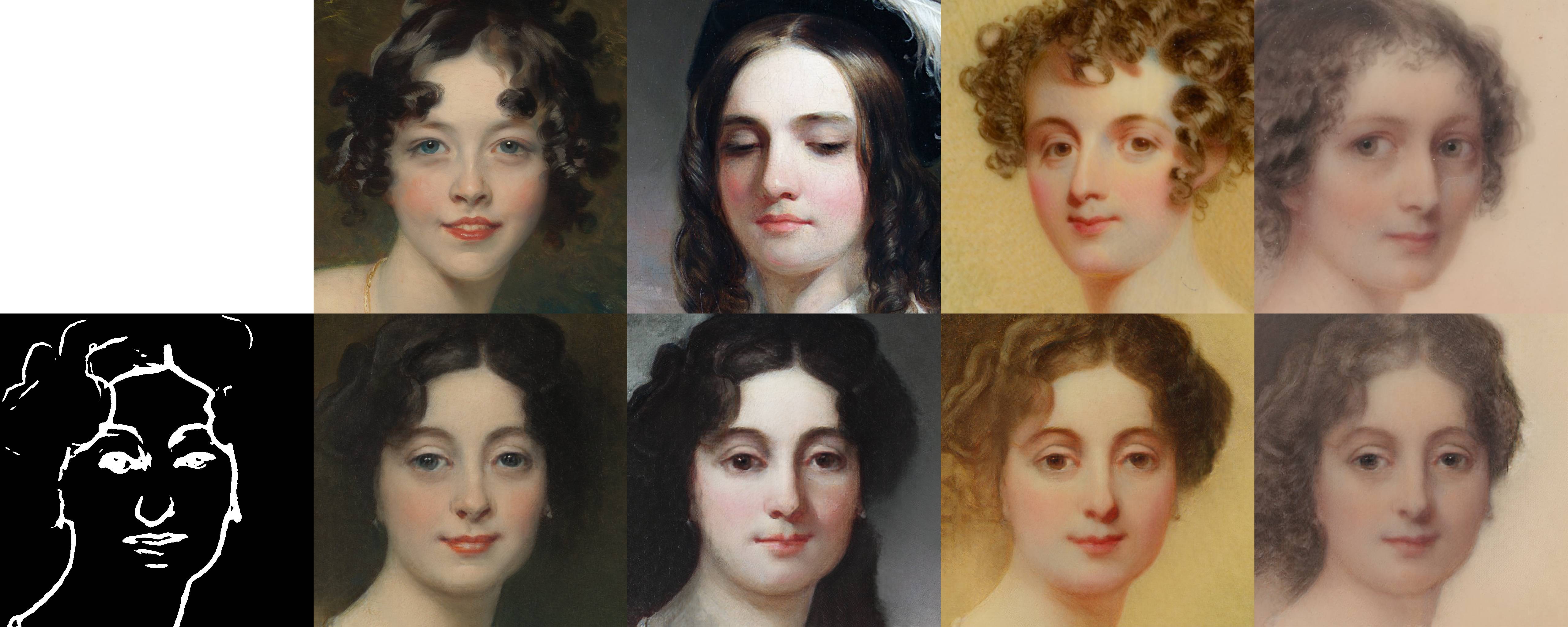}\\
\vspace{+3em}
\includegraphics[width=1\columnwidth]{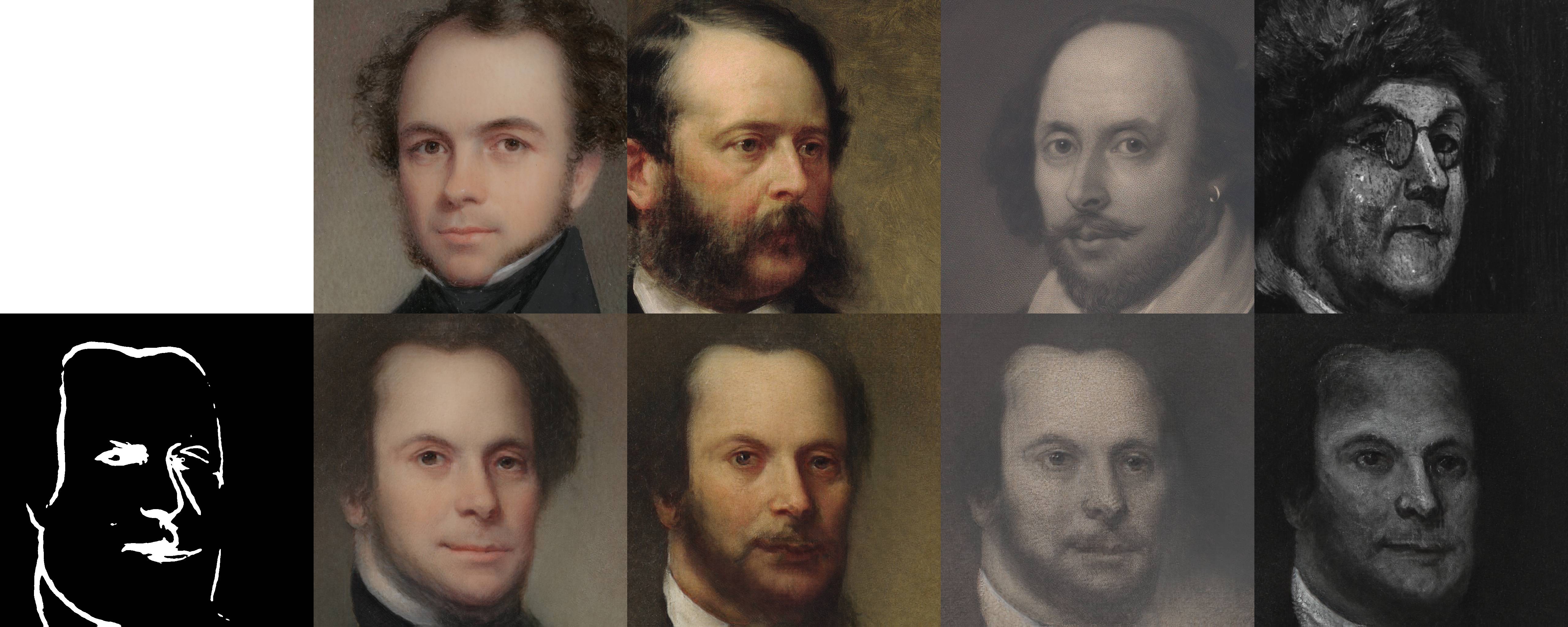}\\
\vspace{+1em}
\caption{Edge-to-face image translation results at resolution $1024\times1024$. 1st row: exemplar images, 2nd row: generated images.  (MetFaces dataset)}
\label{figure:edge_to_face_01}
\end{figure*}

\clearpage
\begin{figure*}[t]
\centering
\includegraphics[width=1\columnwidth]{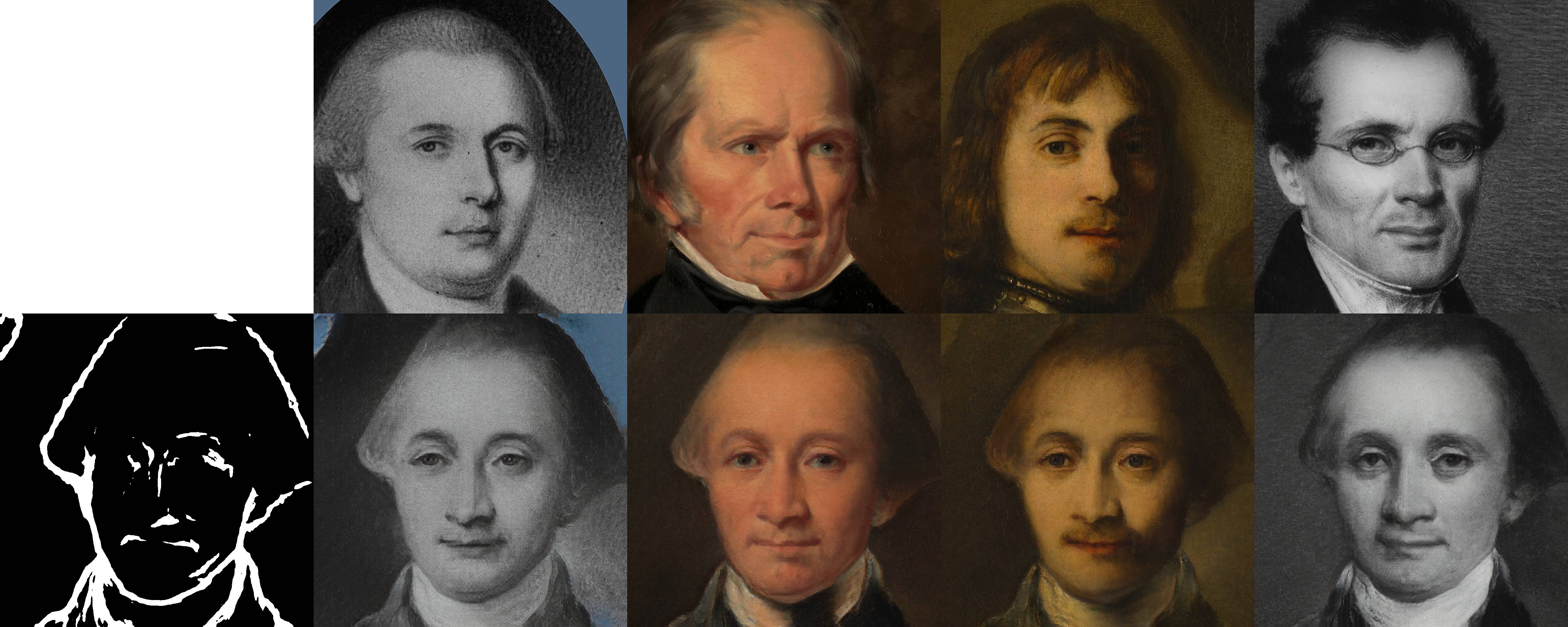}\\
\vspace{+3em}
\includegraphics[width=1\columnwidth]{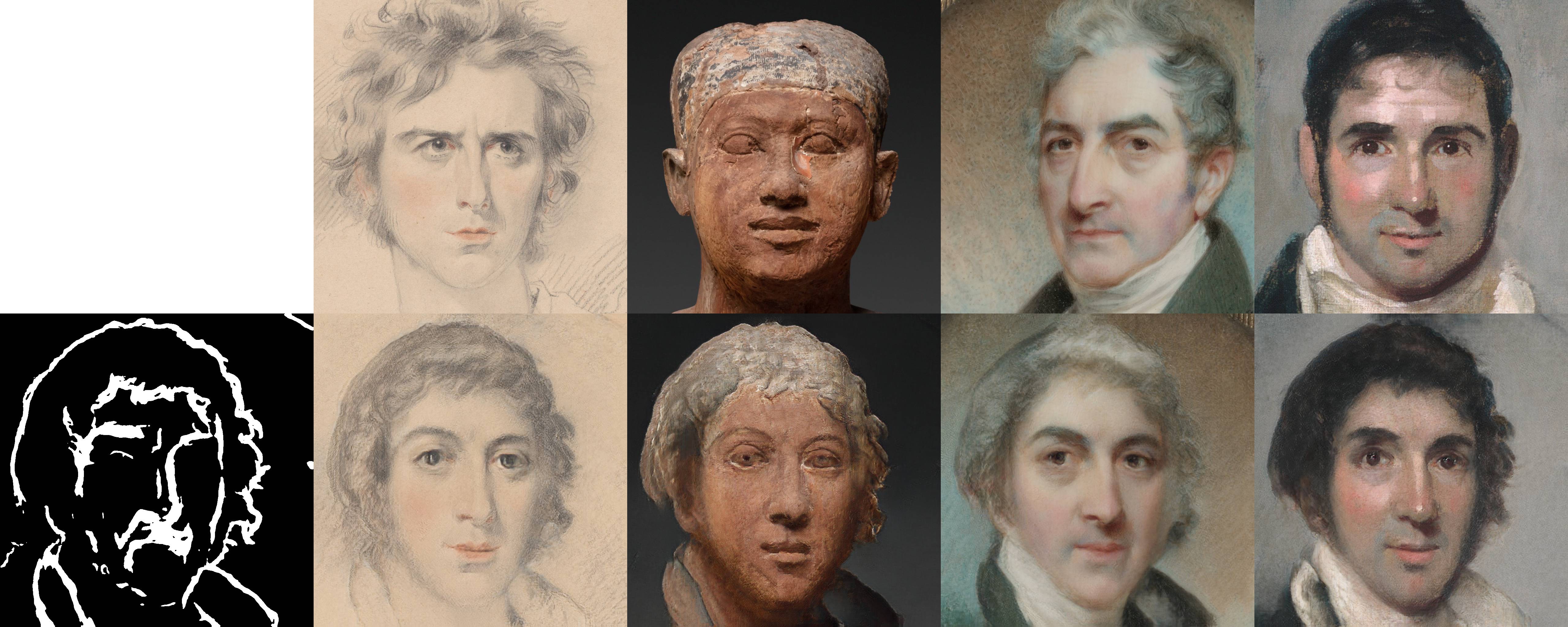}\\
\vspace{+1em}
\caption{Edge-to-face image translation results at resolution $1024\times1024$. 1st row: exemplar images, 2nd row: generated images. (MetFaces dataset)}
\label{figure:edge_to_face_02}
\end{figure*}

\clearpage
\begin{figure*}[t]
\centering
\includegraphics[width=1\columnwidth]{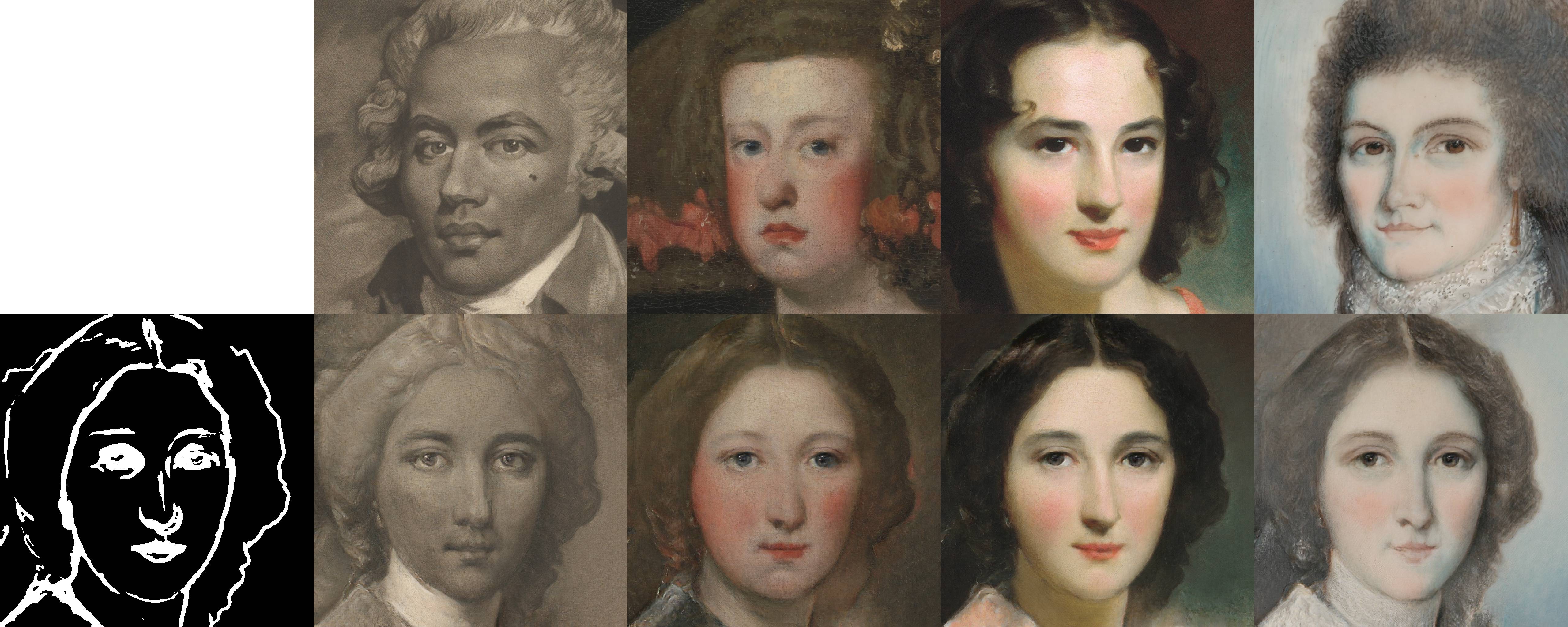}\\
\vspace{+3em}
\includegraphics[width=1\columnwidth]{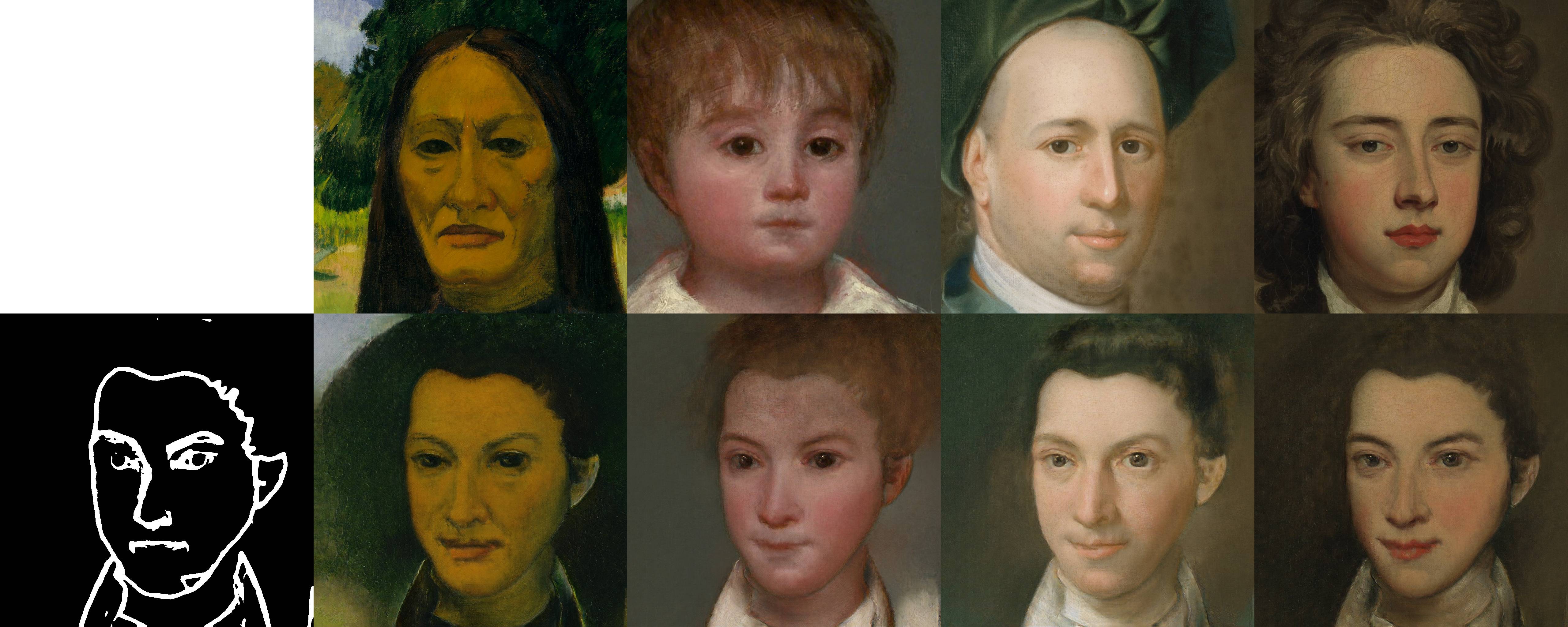}\\
\vspace{+1em}
\caption{Edge-to-face image translation results at resolution $1024\times1024$. 1st row: exemplar images, 2nd row: generated images.  (MetFaces dataset)}
\label{figure:edge_to_face_03}
\end{figure*}

\clearpage
\begin{figure*}[t]
\centering
\includegraphics[width=1\columnwidth]{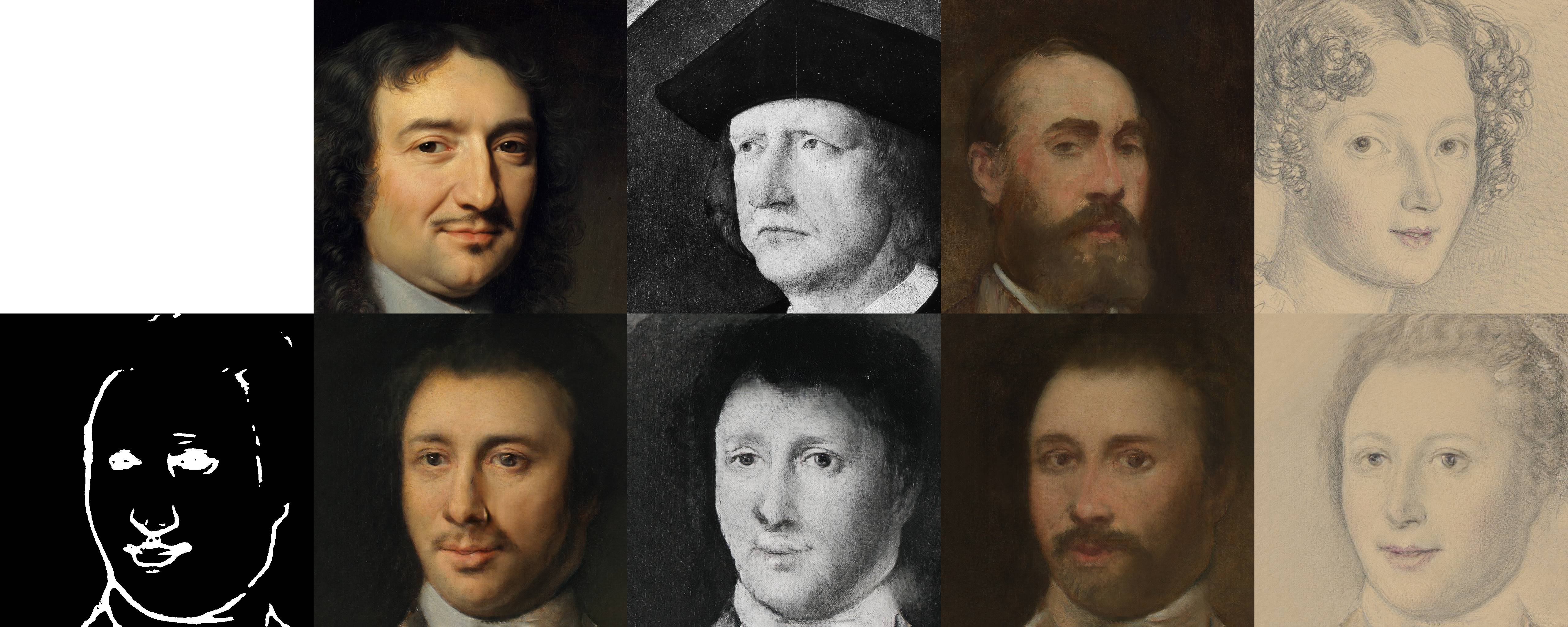}\\
\vspace{+3em}
\includegraphics[width=1\columnwidth]{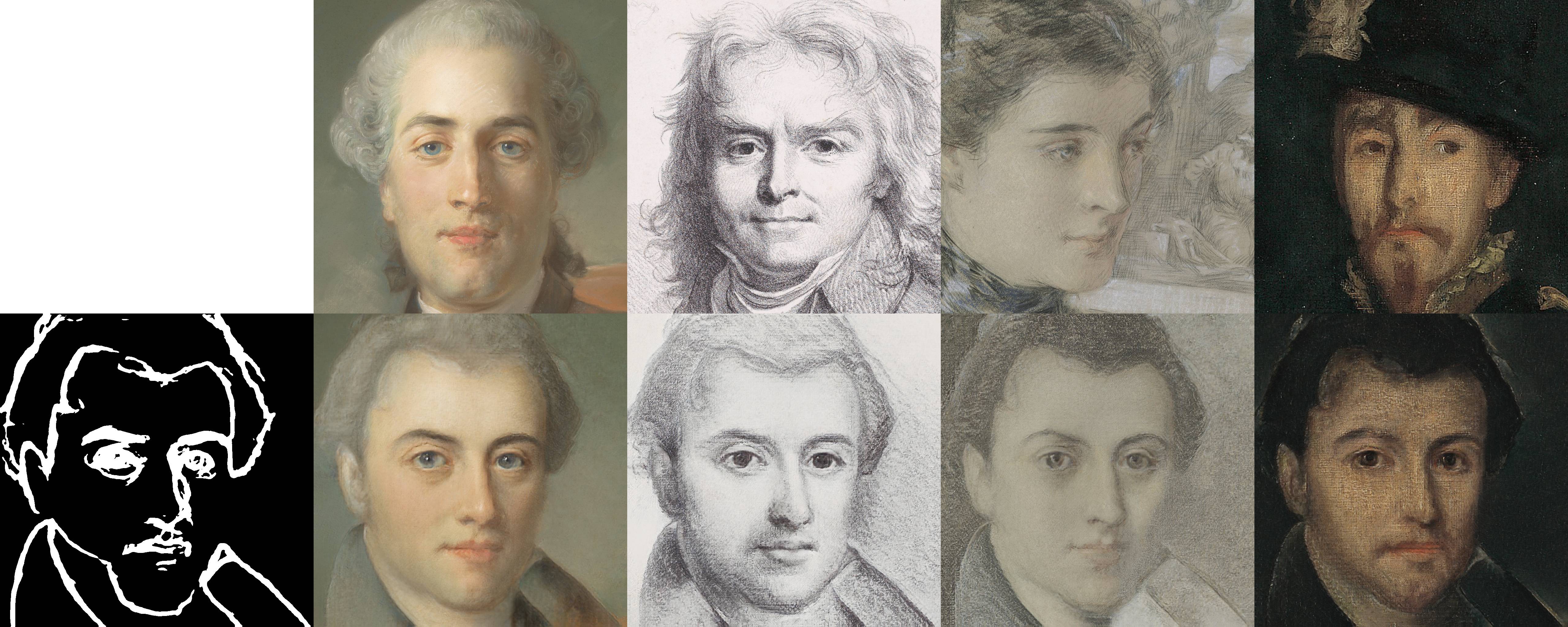}\\
\vspace{+1em}
\caption{Edge-to-face image translation results at resolution $1024\times1024$. 1st row: exemplar images, 2nd row: generated images.  (MetFaces dataset)}
\label{figure:edge_to_face_04}
\end{figure*}

\clearpage
\subsection{Mask-to-image}
Figure~\ref{figure:mask_to_image_01} shows more results of mask-to-image generation on the ADE20K dataset. The proposed method is able to achieve state-of-the-art quality for diverse scenes on this challenging dataset.

\vspace{+2em}
\begin{figure*}[hbtp]
\centering

\begin{tabularx}{0.98\columnwidth}{YYYYYY}
Mask & Our result & Exemplar & Mask & Our result & Exemplar
\end{tabularx}
\renewcommand{\arraystretch}{0.0}

\includegraphics[width=0.49\columnwidth]{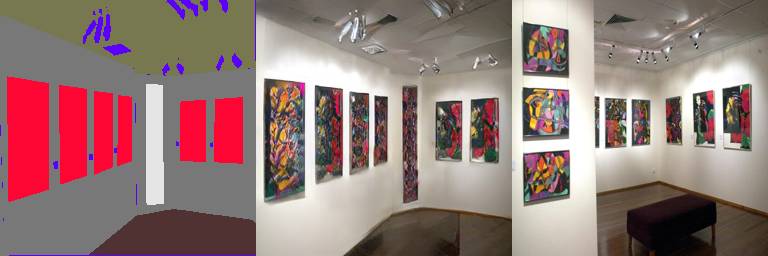}
\includegraphics[width=0.49\columnwidth]{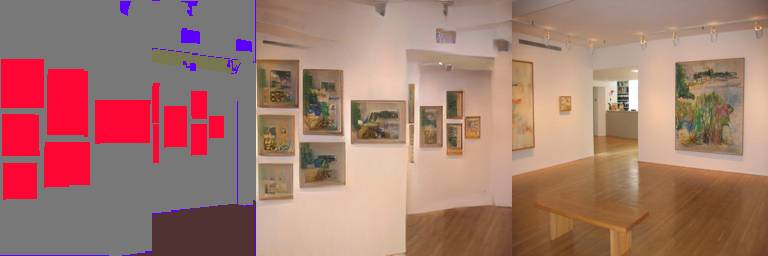}\\
\vspace{+1em}

\includegraphics[width=0.49\columnwidth]{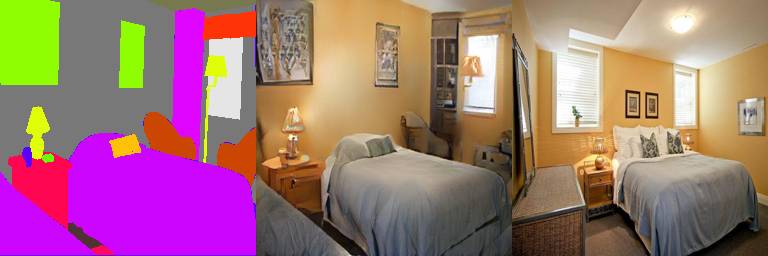}
\includegraphics[width=0.49\columnwidth]{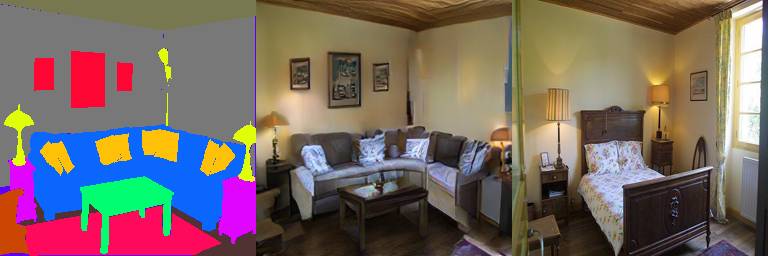}\\
\vspace{+1em}

\includegraphics[width=0.49\columnwidth]{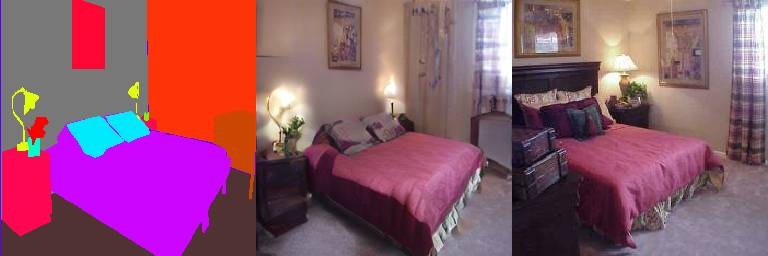}
\includegraphics[width=0.49\columnwidth]{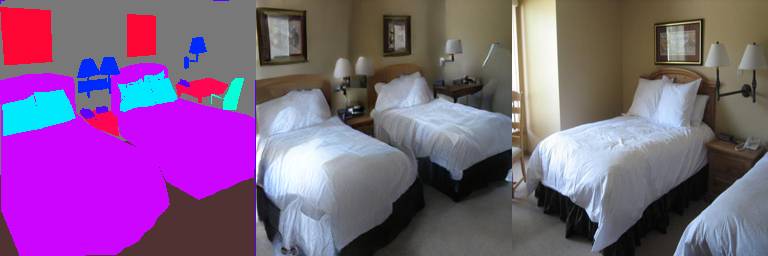}\\
\vspace{+1em}

\includegraphics[width=0.49\columnwidth]{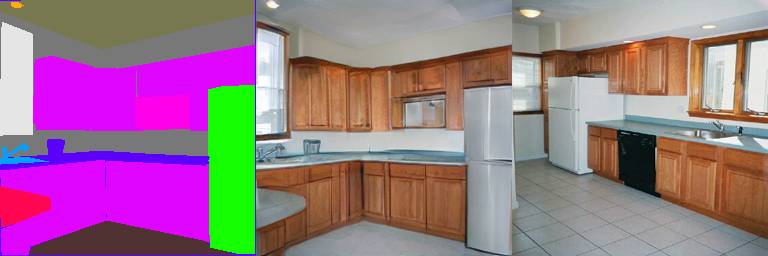}
\includegraphics[width=0.49\columnwidth]{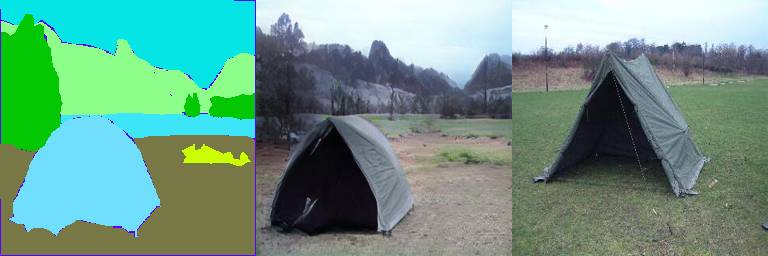}\\
\vspace{+1em}

\includegraphics[width=0.49\columnwidth]{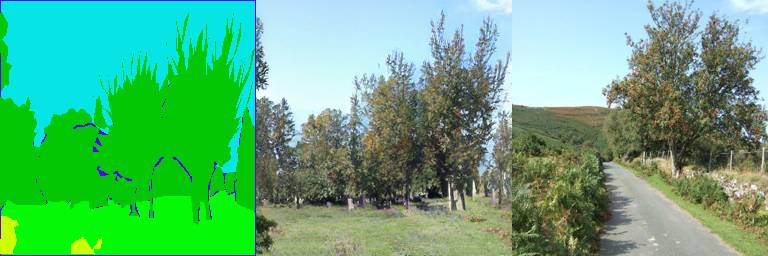}
\includegraphics[width=0.49\columnwidth]{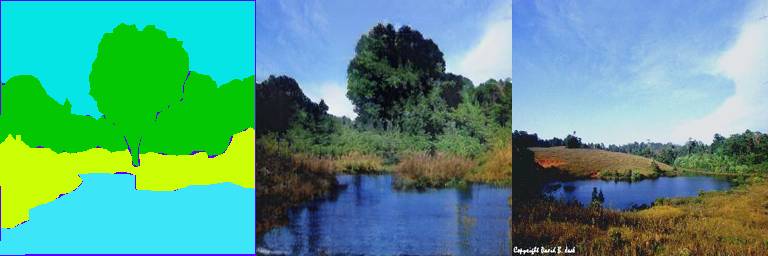}\\

\vspace{+1em}
\caption{Mask-to-image generation results. (ADE20K dataset)}
\label{figure:mask_to_image_01}
\end{figure*}

\clearpage
\subsection{Oil portrait}
Figure~\ref{figure:oil_portrait_01} to Figure~\ref{figure:oil_portrait_02} show more results of oil portrait. Our method takes the edge from real people (CelebA dataset) as input. The output looks like transferring the real person into the oil painting. While the model is purely trained using the paintings in the MetFaces dataset, the model could generalize well to the sketches of real faces.  

\vspace{+3em}
\begin{figure*}[hbtp]
\centering
\includegraphics[width=1\columnwidth]{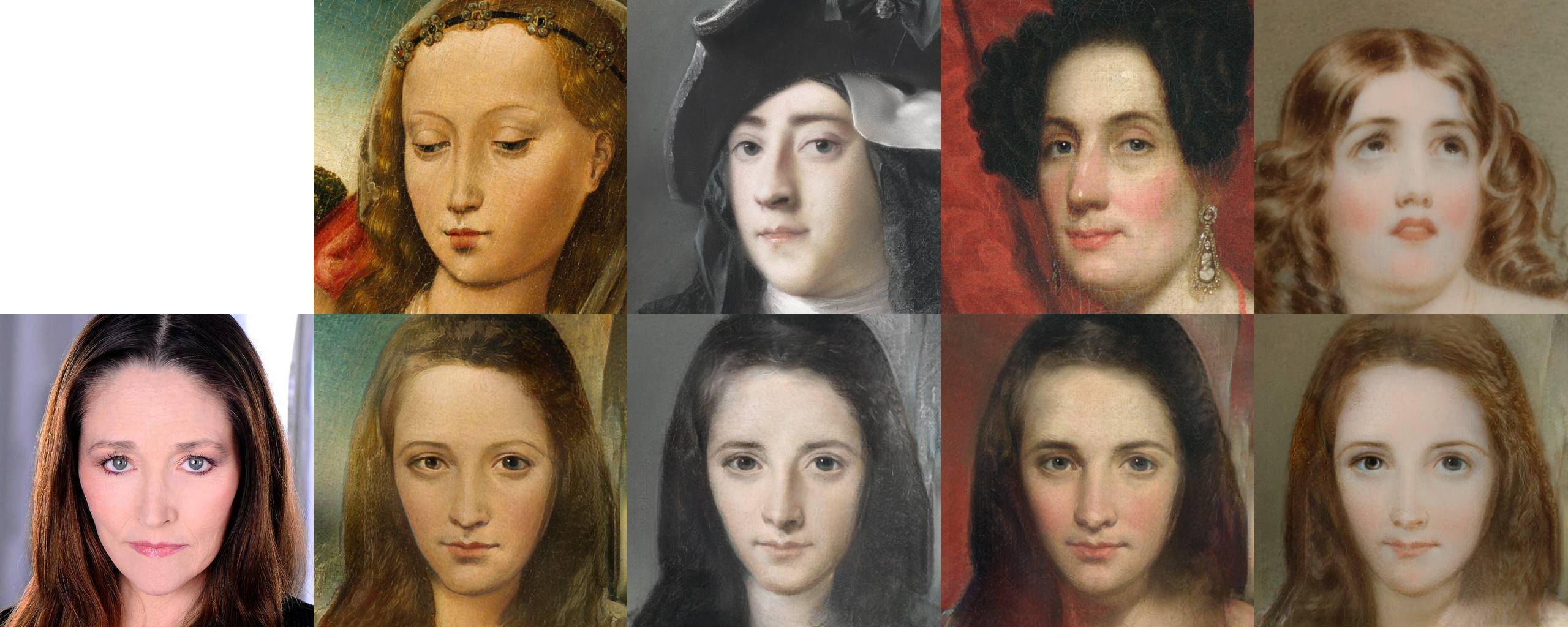}\\
\vspace{+3em}
\includegraphics[width=1\columnwidth]{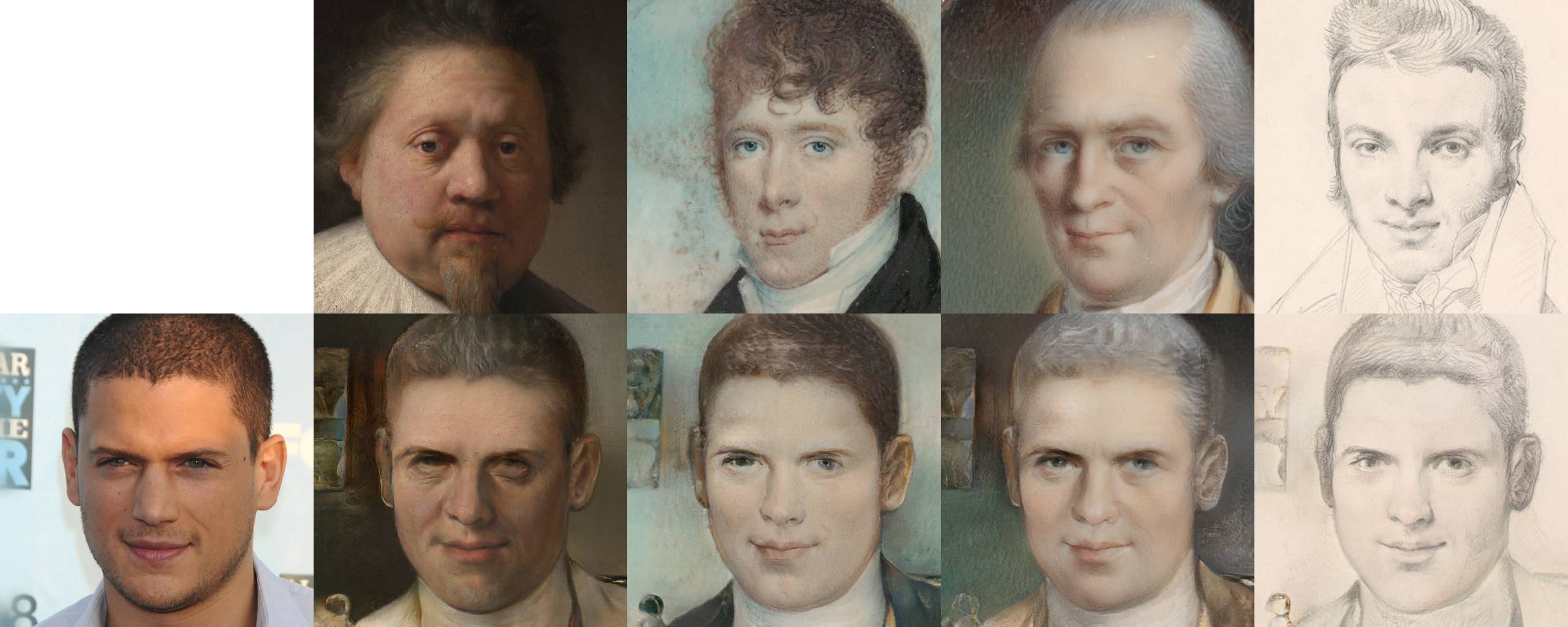}\\
\vspace{+1em}
\caption{Oil portrait results with resolution $512\times 512$. The edge is from the CelebA dataset while the exemplar is from the MetFaces dataset. 1st row: exemplar images, 2nd row: generated images.}
\label{figure:oil_portrait_01}
\end{figure*}

\begin{figure*}[t]
\centering
\includegraphics[width=1\columnwidth]{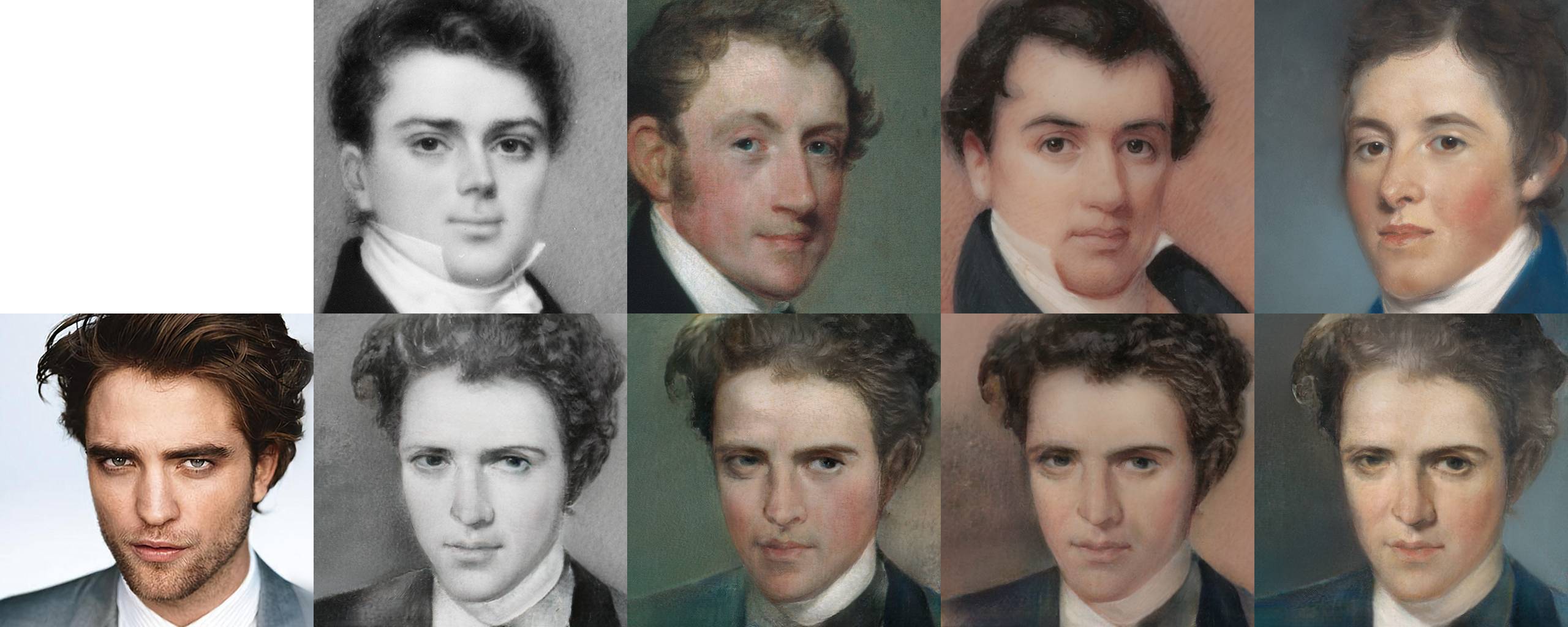}\\
\vspace{+3em}
\includegraphics[width=1\columnwidth]{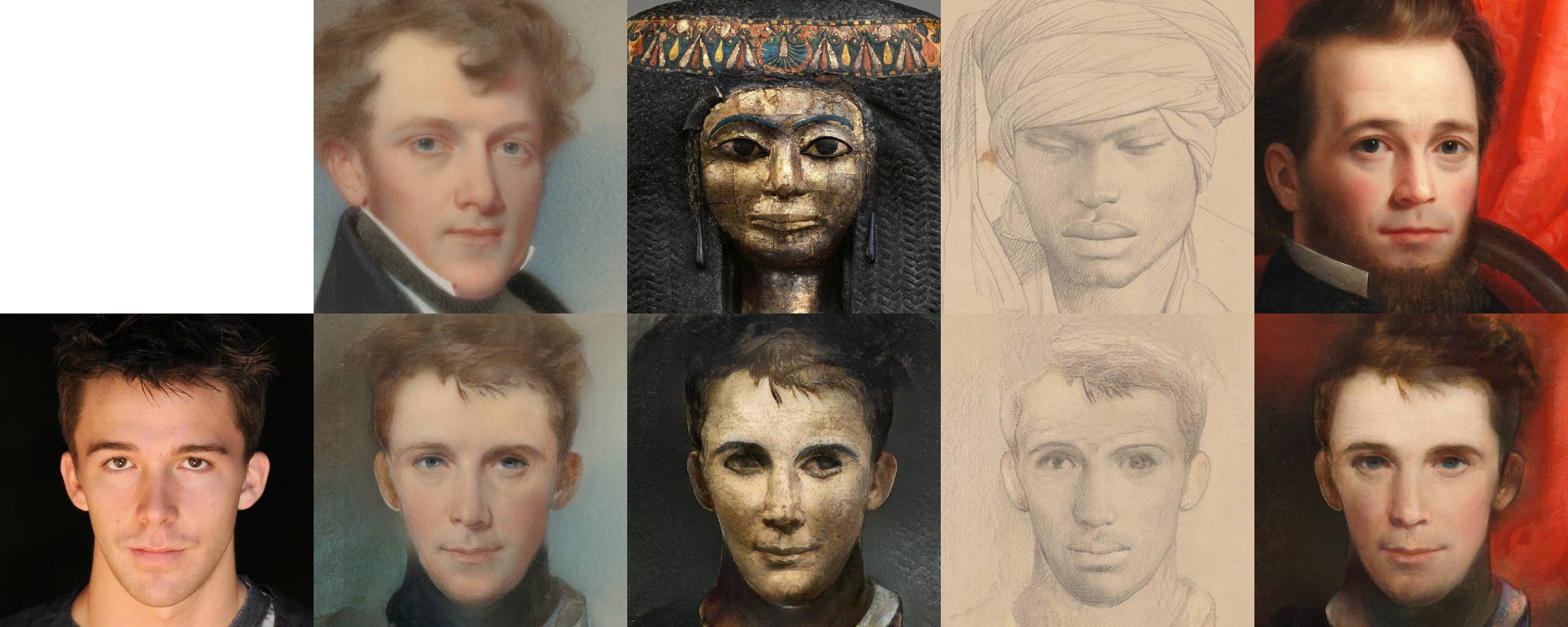}\\
\vspace{+1em}
\caption{Oil portrait results with resolution $512\times 512$. The edge is from the CelebA dataset while the exemplar is from the MetFaces dataset. 1st row: exemplar images, 2nd row: generated images.}
\label{figure:oil_portrait_02}
\end{figure*}

\clearpage
\section{Implementation Details}
Our Hierarchical GRU-assisted PatchMatch establishes full-correspondence with multi-level features. We take the generation resolution $512\times512$ as an example to elaborate upon the implementation details. We choose $L=4$ levels for the resolution $512\times512$ translation, so we establish correspondence on the $64\times64$, $128\times128$, $256\times256$, and $512\times512$ levels.

\vspace{+0.5em}
\noindent \textbf{Hierarchical strategy. }Our method establishes the correspondences via the hierarchical strategy. The smallest scale that we use in the experiments is $64\times64$. Please note that we calculate all the pair-wise similarities on this scale, \ie, we make the two features ($\mathbf{f}^x_{1}$ and $\mathbf{f}^y_{1}$) flatten and calculate the similarity matrix on this scale. We do not rely on sparse matching and spatial propagation at this scale because the correspondence learning is guided by the warped images -- in an indirect manner rather than providing the correspondence ground-truth, and it is difficult to use sparse matching to establish reliable correspondence when the features are not well-learned and appear noisy at the early training phase.

\vspace{+0.5em}
\noindent \textbf{GRU-assisted PatchMatch. }The GRU-assisted PatchMatch module requires the local correspondences in the subsequent higher-scale level, \ie, the scale of $128\times128$, $256\times256$, and $512\times512$. We choose $K=16$ nearest neighbors as candidates for each feature point and the PatchMatch is differentiable as we compute the soft matching by averaging across all these matchings and gradient can be back-forwarded to multiple locations.

\vspace{+0.5em}
\noindent \textbf{Translation network. }The translation network takes the warped exemplar images of multi-levels as input and synthesizes the final output according to the exemplar style. The warped exemplar images of multi-levels are first resized to the same scale ($512\times512$ in this example) and then concatenated along the channel dimension. Two convolutional layers digest this concatenation input and produce the parameters for style modulation. We use \emph{positional normalization}~\cite{li2019positional} within this sub-network, with the denormalization modulated by the warped exemplar.

\vspace{+0.5em}
\noindent \textbf{The detailed architecture. }Table~\ref{table:architecture} shows the implementation details of our method, with the naming convention as the CycleGAN~\cite{zhu2017unpaired}. Please note we take the generation at the resolution $512\times512$ as an example, and the network can be adapted to even higher resolutions.

\vspace{+0.5em}
\noindent \textbf{Geometric data augmentation. }The geometric augmentation $\mathcal{T}$ includes flip, random crop, and piecewise affine transformation, which is used to form the pseudo exemplar $\mathcal{T}(x_A)$. The training triplet thus becomes: input $x_A$, pseudo exemplar $\mathcal{T}(x_A)$ and the desired output $x_B$.

\vspace{+0.5em}
\noindent \textbf{The protocol of sampling examples. }1) For training: we randomly sample images of the same person but of a different pose as exemplars for DeepFashion whereas example oil portraits are randomly sampled for MetFaces; for ADE20k, we retrieve the top 20 images in terms of pretrained VGG feature distance. 2) For figures reported in the paper: we randomly sample 10 exemplars for DeepFashion and MetFaces, and retrieve top-5 exemplars for ADE20k.

\vspace{+0.5em}
\noindent \textbf{The speed. }It takes around 200h to train 100 epochs using 8 NVIDIA V100 GPUs. During inference, it takes $\sim$0.6s to synthesize an $512\times512$ image. This information will be added in the final paper.

\vspace{+2em}
\renewcommand{\arraystretch}{1.1}
\begin{table}[!tbh]
\small
\centering
\begin{tabular}{@{}l|l|l|l@{}}
\toprule
{Sub-network} & {Module} & {Layers in the module} & {Output shape (H$\times$W$\times$C)}\\
\midrule
\multirow{7}{2.5cm}{Multi-level Domain Alignment Network} 
& \multirow{6}{3.0cm}{Adaptive Domain Feature Encoder$\times2$} 
& Conv2d / k3s1 + Resblock / k3s1 & 512$\times$512$\times$64\\ 
& & Conv2d / k4s2 + Resblock / k3s1& 256$\times$256$\times$128\\ 
& & Conv2d / k4s2 + Resblock / k3s1& 128$\times$128$\times$256\\ 
& & Conv2d / k4s2 + Resblock / k3s1& 64$\times$64$\times$512\\ 
& & Bilinear Interpolation + Resblock / k3s1 & 128$\times$128$\times$256\\ 
& & Bilinear Interpolation + Resblock / k3s1 & 256$\times$256$\times$128\\ 
& & Bilinear Interpolation + Resblock / k3s1 & 512$\times$512$\times$64\\ 
\cmidrule{2-4}
&Correspondence &(GRU-assisted PatchMatch \& Warping) $\times4$ & 64$\times$64$\times$3\\
& &  & 128$\times$128$\times$3\\
& &  & 256$\times$256$\times$3\\
& &  & 512$\times$512$\times$3\\
\midrule
\multirow{7}{2.5cm}{Translation Network}
& \multirow{3}{3.0cm}{Style Encoder$\times7$} 
& Bilinear Interpolation 
& $h^i \times w^i \times$3\\ 
& & Conv2d / k3s1 & $h^i \times w^i \times$128\\
& & Conv2d / k3s1 & $h^i \times w^i \times c^i$\\
\cmidrule{2-4}
& \multirow{3}{3.0cm}{Generator} 
& Conv2d / k3s1 & 8$\times$8$\times$1024\\ 
&  & Resblock$\times7$ & 256$\times$256$\times$64\\
&  & Conv2d / k3s1 & 256$\times$256$\times$3\\
\bottomrule
\end{tabular}
\vspace{+0.5em}
\caption{{The detailed architecture of our approach. k3s1 indicates the convolutional layer with kernel size 3 and stride 1.}
}
\label{table:architecture}
\end{table}
\vspace{-1em}

%-------------------------------------------------------------------------

\end{onecolumn}

%-------------------------------------------------------------------------
\clearpage
\begin{twocolumn}
{\small
\bibliographystyle{ieee_fullname}
\bibliography{egbib}
}
\end{twocolumn}

\end{document}